\pdfoutput=1

\documentclass[11pt]{article}

\usepackage[final]{acl}

\usepackage{times}
\usepackage{latexsym}
\usepackage{multirow}

\usepackage[T1]{fontenc}

\usepackage[utf8]{inputenc}

\usepackage{microtype}

\usepackage{inconsolata}

\usepackage{graphicx}
\usepackage{amsmath}
\usepackage{nicematrix}
\usepackage{booktabs}
\usepackage{pdflscape}
\usepackage{hyperref}

\newcommand{\approxtext}[1]{\ensuremath{\stackrel{\text{#1}}{\approx}}}
\title{Multimodal Causal Reasoning Benchmark: Challenging Multimodal Large Language Models to Discern Causal Links Across Modalities}

\author{
 \textbf{Zhiyuan Li\textsuperscript{}},
 \textbf{Heng Wang\textsuperscript{}},
 \textbf{Dongnan Liu\textsuperscript{}},
 \textbf{Chaoyi Zhang\textsuperscript{}},
\\
 \textbf{Ao Ma\textsuperscript{}},
 \textbf{Jieting Long\textsuperscript{}},
 \textbf{Weidong Cai\textsuperscript{}}
\\
 \textsuperscript{}School of Computer Science, The University of Sydney
\\
 \small{\{zhli0736, hwan9147, czha5168, aoma0081, jlon5443\}@uni.sydney.edu.au 
 }
 \\
 \small{
 \{dongnan.liu, tom.cai\}@sydney.edu.au
 }
}

\definecolor{rowgray}{HTML}{EFEFEF}
\definecolor{darkgray}{HTML}{D1D1D1}
\begin{document}
\maketitle
\begin{abstract}
Multimodal Large Language Models (MLLMs) have showcased exceptional Chain-of-Thought (CoT) reasoning ability in complex textual inference tasks including causal reasoning. However, will these causalities remain straightforward when crucial hints hide in visual details? If not, what factors might influence cross-modal generalization? Whether we can effectively enhance their capacity for robust causal inference across both text and vision? Motivated by these, we introduce \textbf{MuCR} - a novel \textbf{Mu}ltimodal \textbf{C}ausal \textbf{R}easoning benchmark that leverages synthetic siamese images and text pairs to challenge MLLMs. Additionally, we develop tailored metrics from multiple perspectives, including image-level match, phrase-level understanding, and sentence-level explanation, to comprehensively assess MLLMs' comprehension abilities. Our experiments reveal that current MLLMs fall short in multimodal causal reasoning compared to their performance in purely textual settings. Additionally, we find that identifying visual cues across images is key to effective cross-modal generalization. Finally, we propose a \textbf{VcCoT} strategy that better highlights visual cues, and our results confirm its efficacy in enhancing multimodal causal reasoning.
The project is available at: \url{https://github.com/Zhiyuan-Li-John/MuCR}
\end{abstract}

\section{Introduction}
\begin{figure}[!ht]
  \centering
  \vspace{-0.0cm}   \includegraphics[width=1.0\linewidth]{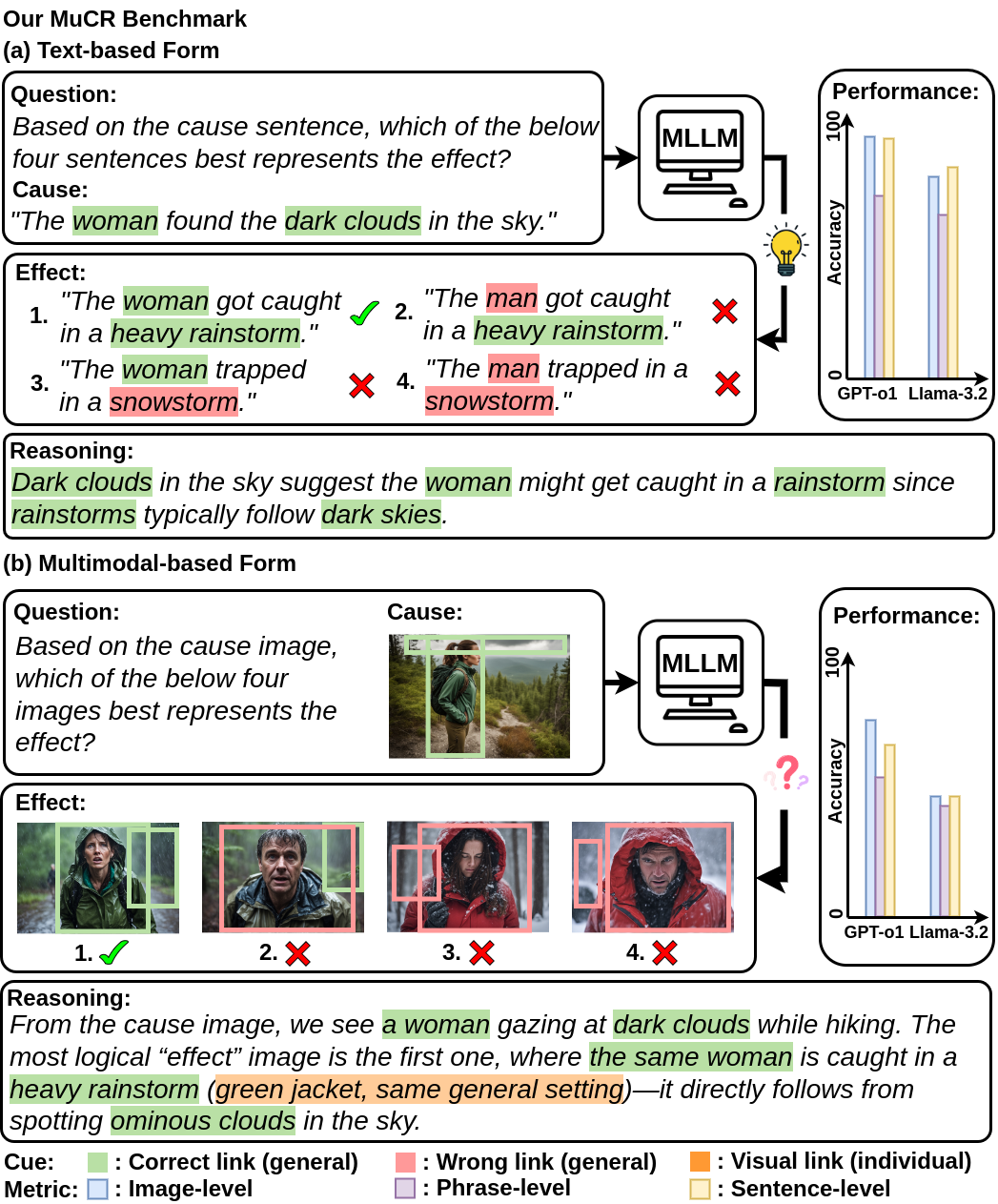}
   \vspace{-0.4cm}
   \caption{An example from MuCR challenges MLLMs with weather-related causality across two modalities.}
   \label{fig1}
   \vspace{-0.4cm}
\end{figure}

\begin{figure*}[t]
  \centering
  \vspace{-0.5cm}   \includegraphics[width=0.98\linewidth]{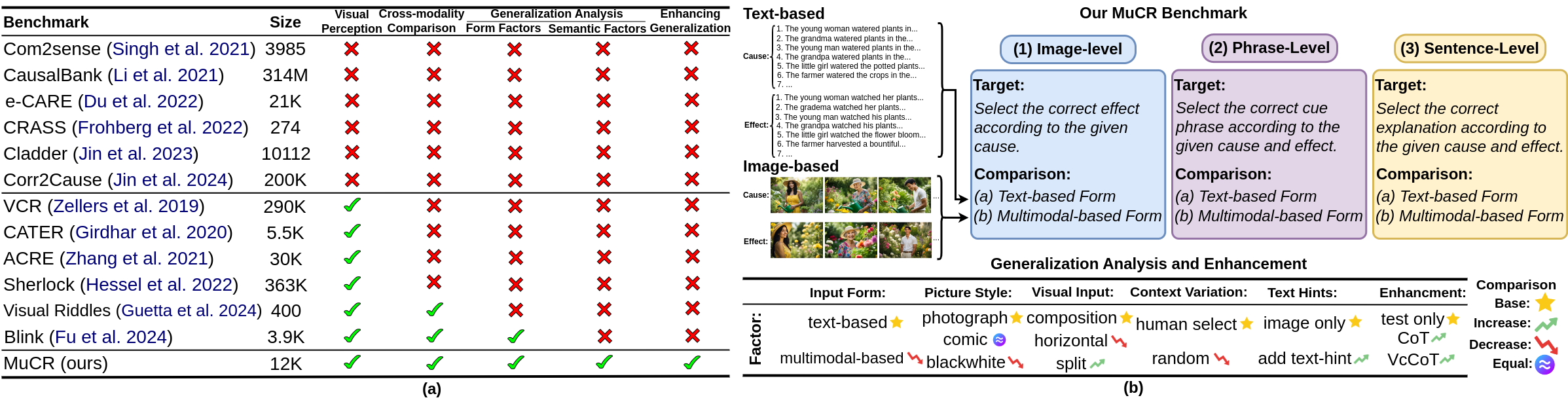}
   \vspace{-0.0cm}
   \caption{(a) Comparison of our MuCR and related datasets on reasoning tasks. (b) Detailed illustration of our dataset structure and corresponding cross-modal generalization exploration.}
   \label{fig2}
   \vspace{-0.3cm}
\end{figure*}

Causal reasoning is the process of identifying the relationship between a cause and its effect, which is regarded as a fundamental capability of artificial intelligence~\cite{liu2024large}. Recent advancements in CoT reasoning capabilities of MLLMs~\cite{gpto1, guo2025deepseek} have driven significant progress in complex analytical tasks, including causal reasoning within the textual modality~\cite{jin2023cladder, bagheri2024c2p, ashwani2024cause}. These developments involve enabling MLLMs to generate coherent explanations~\cite{kiciman2023causal}, providing multi-step chain-of-thought (CoT)~\cite{bao2024llms}, or even analyzing complex causal relationships that typically demand expert-level topological structure knowledge~\cite{vashishtha2023causal}. Despite these advancements, existing linguistic benchmarks~\cite{singh2021com2sense, du2022care, jin2023cladder} are beginning to fall short in assessing the more advanced visual capabilities of the latest MLLMs such as GPT-o1~\cite{gpto1}, Deepseek-R1~\cite{guo2025deepseek}, Gemini-1.5~\cite{gemini1.5}, and Claude-3.5~\cite{claude3.5}, not to mention facilitating cross-modal comparison and analysis (as shown in Figure~\ref{fig1}).

Following this, we propose three key questions: Can MLLMs achieve the same level of causal reasoning comprehension as they do in textual modality? If not, what factors might influence cross-modal generalization? How can we enhance their capacity for robust causal inference? We find that most existing benchmarks fail to address such comparisons or support further exploration in this area. Especially, as shown in Figure~\ref{fig2} (a), we identify two major drawbacks in previous benchmarks: \textbf{Absence of visual modality}: Linguistic causal reasoning benchmarks~\cite{singh2021com2sense, li2021guided, du2022care, frohberg2022crass, jin2023cladder, jin2024can} fail to assess visual comprehension ability of MLLMs. \textbf{Incomplete of cross-modal analysis}: Most causal reasoning VQA tasks~\cite{zellers2019recognition, girdhar2020cater, zhang2021acre, hessel2022abduction} neglect cross-modal comparison. Recently, some benchmarks~\cite{bitton2024visual, fu2024blink} have begun exploring this domain. For instance, Blink~\cite{fu2024blink} examines cross-modal comparisons and conducts basic generalization analyses involving factors like shape and size. As illustrated in Figure~\ref{fig2}(b), our proposed MuCR comprehensively evaluates causal reasoning at the image, phrase, and sentence levels and offers a multi-faceted analysis of cross-modal generalization that encompasses both visual form factors and semantic elements. Moreover, we propose a novel VcCoT strategy to further enhance cross-modal generalization by improving visual cue perception.

We evaluate current state-of-the-art (SOTA) MLLMs on our MuCR benchmark. Experiment results indicate that all models fall short of human performance, particularly in multimodal settings. Moreover, they exhibit a pronounced cross-modal gap when discerning causal links across modalities. In addition, we conduct in-depth generalization analysis and demonstrate that visual semantic factors, especially the ability to identify visual cues across siamese images, play a pivotal role.
 
Our contributions are summarized as follows:
\begin{itemize}
    \item We identify the limitations of current causal reasoning benchmarks, including failing to evaluate the advanced visual capabilities of the latest MLLMs and offering incomplete cross-modal analyses.
    \item We propose the MuCR benchmark, which can comprehensively evaluate MLLMs' causal reasoning ability across two modalities.
    \item Our extensive experiments with SOTA MLLMs reveal interesting insights and suggest potential directions for future research.
\end{itemize}

\section{Related Work}
\begin{figure*}[!th]
  \centering
  \vspace{-0.0cm}   \includegraphics[width=1.0\linewidth]{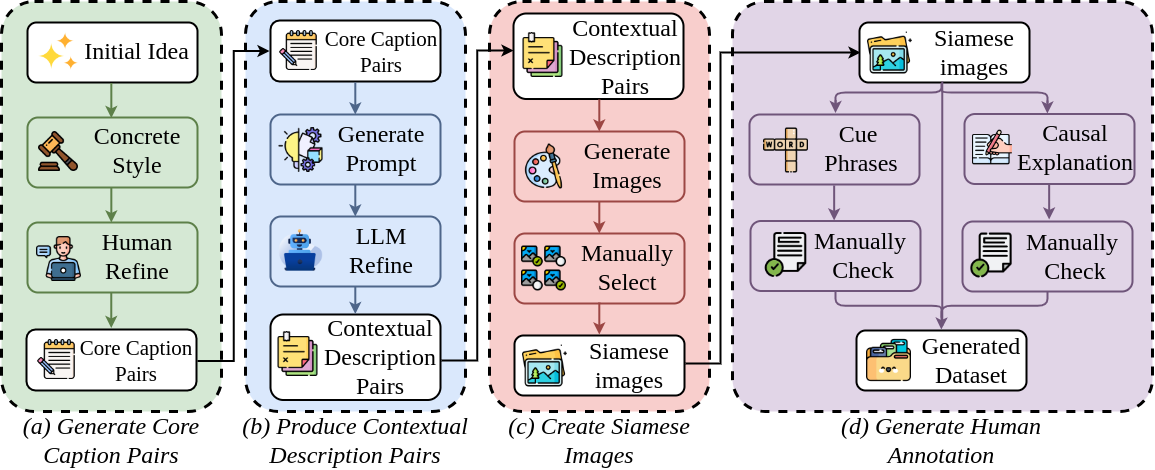}
   \vspace{-0.1cm}
   \caption{The overview of our MuCR benchmark construction process. It follows synthesis in four core levels: generating core caption pairs, producing contextual description pairs, creating siamese images, and generating human annotations.}
   \label{fig3}
   \vspace{-0.1cm}
\end{figure*}
\subsection{Causal Reasoning}
The ability to perform causal reasoning is widely considered a core feature of artificial intelligence. With the development of Large Language Models (LLMs), they have exhibited increasingly robust capabilities in causal reasoning tasks. Previous benchmarks, such as Com2sense~\cite{singh2021com2sense} and CausalBank~\cite{li2021guided}, are becoming insufficient for evaluating linguistic abilities. To address this, \citet{romanou2023crab} introduced the CRAB benchmark, which requires LLMs to capture explicit causal relationships in real-world scenarios. However, these benchmarks focus solely on the text modality, leaving the crucial question of multimodal reasoning unaddressed. \citet{hessel2022abduction} introduced Sherlock to challenge MLLMs in identifying visual clues scattered throughout a scene and making reasoning inferences combined with commonsense and life experience. More recently, \citet{guetta2024visual} and \citet{fu2024blink} presented complex visual reasoning challenges to further explore MLLMs’ capabilities. Although these benchmarks have considered the visual modality, they still fail to comprehensively analyze cross-modal generalization capacity. In this paper, we make an early attempt to extensively explore multimodal causal reasoning tasks across modalities.

\subsection{LLMs' Generalization}
The field of LLMs generalization has gained significant traction in recent years, with numerous tasks proposed to evaluate models’ ability to handle previously unseen contexts and domains. Existing tasks can be broadly divided into compositional, cross-task, cross-lingual, cross-domain, and robustness-based categories. Compositional tasks, such as CFQ~\cite{keysers2020measuring} and COGS~\cite{kim2020cogs}, test whether models can systematically combine smaller linguistic units to form novel expressions. Cross-task generalization often involves multi-task learning setups, such as DecaNLP~\cite{mccann2018natural} and BIG-Bench~\cite{srivastava2022beyond}, where models must adapt to tasks with minimal guidance. Cross-lingual benchmarks, like XNLI~\cite{conneau2018xnli} and XTREME~\cite{hu2020xtreme}, measure performance across languages, while cross-domain tasks emphasize shifting between specialized fields~\cite{li2023llm,zhou2024geng}.
Meanwhile, robustness-oriented evaluations such as HellaSwag~\cite{zellers2019hellaswag} and adversarial GLUE~\cite{wang2021adversarial} assess how well models withstand noisy, ambiguous, or adversarial inputs. In this paper, we shift our focus to the generalization in multimodal causal reasoning tasks, conducting a concise but comprehensive analysis of the factors that hinder cross-modal generalization and exploring strategies to enhance it for robust causal reasoning.

\begin{figure*}[!t]
  \centering
  \vspace{-0.7cm}   \includegraphics[width=1.0\linewidth]{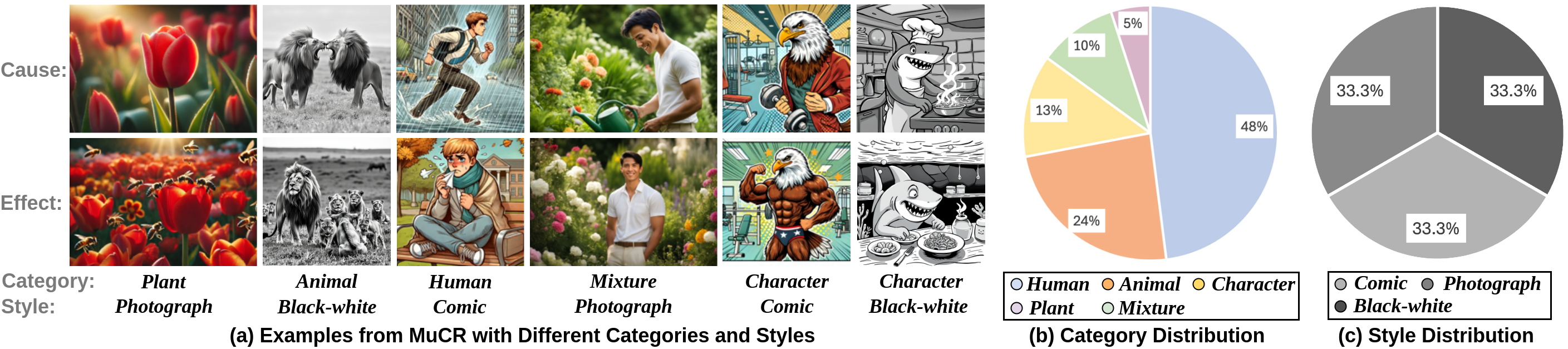}
   \vspace{-0.3cm}
   \caption{(a) Examples from our MuCR dataset featuring different categories and styles. The ``Mixture" category represents two or more tags involved in the causality. (b) Category distribution overview showing the proportions of human, animal, character, plant, and mixture categories. (c) Style distribution overview illustrating the proportions of comic, photographic, and black-white styles.}
   \label{fig4}
   \vspace{-0.3cm}
\end{figure*}

\section{The MuCR Dataset}
\label{M3.0}
In this section, we detail the construction of the MuCR dataset. Figure~\ref{fig3} illustrates the systematic workflow of our multimodal cause-and-effect benchmark generation including: generating core caption pairs, producing contextual description pairs, creating siamese images, and generating human annotations (see Appendix~\ref{A1.0} for further examples and details).

\subsection{Dataset Creation}
\paragraph{Generating Core Caption Pairs.}
The MuCR benchmark is designed to assess MLLMs’ ability to perform causal inference across modalities. To achieve this, we begin by generating core caption pairs that clearly illustrate cause-and-effect relationships. In order to minimize individual bias, we employ twelve volunteers and group each two as a team: one processes and refines the captions based on initial ideas and iterative feedback, while the other reviews them and offers suggestions for improvement (see Appendix~\ref{A1.1} for an explanation of why we structure the generation process this way, as well as illustrative examples). Through these steps, we create 4,000 cause-and-effect caption pairs.

\paragraph{Producing Contextual Description Pairs.}
While core caption pairs effectively depict the cause-and-effect relationship, they often lack contextual details such as appearance, clothing, and environmental context that serve as crucial visual cues for high-quality cause-and-effect image synthesis. To address this issue, we leverage the linguistic capabilities of LLMs to enhance core caption pairs by enriching contextual details. By maintaining these elements consistently across images, our approach not only effectively depicts causality at a semantic level but also improves visual coherence (see Appendix~\ref{A1.2} for further explanation).

\paragraph{Create Siamese Images.}
\label{M3.1}
We employ diffusion models with contextual descriptions as prompts to generate cause-and-effect image pairs. Specifically, we utilize DALL-E~\cite{ramesh2021dalle}, DeepAI~\cite{deepai}, Stability-AI~\cite{stabilityai}, and Flux1~\cite{flux.1} for image synthesis, aiming to minimize model bias and enhance the diversity of the generated images. We also incorporate three styles (photograph, comic, and black-white) when creating these images. Specifically, each sentence yields 10 images per style, resulting in 20 images for every cause-and-effect pair in one style (a total of 240k images). Then, volunteers manually select the two representations that best capture the semantic causality and maintain visual consistency. This process produces 12k cause-and-effect image pairs spanning various categories (humans, animals, plants, characters, and mixtures) and three styles (photograph, comic, and black-white). Figure~\ref{fig4} illustrates examples from our MuCR benchmark, showcasing multiple categories and styles alongside an overview of their distribution (see Appendix~\ref{A1.3} for more high-quality samples).

\paragraph{Generate Human Annotation.}
We require volunteers to create text annotations for each cause-and-effect image pair. As shown in Figure~\ref{fig3}, it consists of a phrase-level list (cue phrases) and sentence-level description (cause-and-effect explanations). The cue phrases comprise a list of four options, each being a word or phrase. Among these, only one phrase correctly explains or is highly relevant to the causality, while the other three are striking elements in the images but do not serve as proper cues. The sentence-level annotation is designed to verify whether the MLLMs truly understand multimodal causality and can select reasonable explanations. To achieve this, we require volunteers to structure the explanation by first describing the content of the cause, followed by the content of the effect, and concluding with the causal link connecting between them.

\subsection{Evaluation Metrics}
\paragraph{Image-level Metric.}
The image-level metric is call \textbf{c}ause-to-\textbf{e}ffect (\textbf{C2E}) score. It is designed to assess whether the MLLMs can identify cue links and make the correct choice from four potential effects according to the given cause. Given the cause in the form $\mathcal{G}^*(c)$ (* can either be $\mathcal{G}^{text}$ representing text-based form or $\mathcal{G}^{multi}$ representing multimodal-based form), the model is required to select the optimal choice among four potential effects $\{\mathcal{G}^*(e)^{(i)}\}^{4}_{i=1}$. The \textbf{C2E} score can be computed as follows:

\begin{gather}
S^{*}_{I} = F(Q_{I}, \mathcal{G}^*(c), \{\mathcal{G}^*(e)^{(i)}\}^{4}_{i=1}), \\
f_I(S^{*}_{I}) = \begin{cases}
1, &  S^{*}_{I} = {S^{*}_{I}}'\\
0, & \text{otherwise}
\end{cases}
\end{gather}
\normalsize where $S^{*}_{I}$ represents the MLLMs' prediction. $F$ represents MLLM. $Q_{I}$ represents corresponding question for Image-level. $f_I$ represents the function to calculate the \textbf{C2E} score. ${S^{*}_{I}}'$ represents the correct answer.

\paragraph{Phrase-level Metric.} 
The phrase-level metric is called \textbf{CP} score (\textbf{C}ue \textbf{P}hrase), which tests MLLMs' capability to distinguish the correct cue from a list of fraudulent phrases according to the cause and effect. Given the cause-and-effect pairs $\{\mathcal{G}^*(c), \mathcal{G}^*(e)\}$, the model is required to select the optimal choice among four potential cue phrases $\{T_P^{(i)}\}^{4}_{i=1}$. The \textbf{CP} score can be computed as follows:
\begin{gather}
S^{*}_{P} = F(Q_{P}, \mathcal{G}^*(c), \mathcal{G}^*(e), \{T_P^{(i)}\}^{4}_{i=1})\\
f_P(S^{*}_{P}) = \begin{cases}
1, & S^{*}_{P} = {S^{*}_{P}}'\\
0, & \text{otherwise}
\end{cases}
\end{gather}
 where $S^{*}_{P}$ represents the MLLMs' prediction. $F$ represents MLLM. $Q_{P}$ represents corresponding question for Phrase-level. $f_P$ represents the function to calculate the \textbf{CP} score. ${S^{*}_{P}}'$ represents the correct answer.

\paragraph{Sentence-level Metric.}
Our final metric is designed to evaluate MLLMs’ ability to
identify the correct explanation according to the cause and effect. The sentence-level metric is called the \textbf{exp}lanation (\textbf{EXP}) score. Specifically, we collect four candidate explanations that share similar causalities but differ in their cues. Only one explanation accurately captures the causal relationship and matches the detailed cues, while the other three do not. Given the condition $\{\mathcal{G}^*(c), \mathcal{G}^*(e)\}$ with the corresponding question $Q_S$, the model is required to select the optimal choice among four potential explanations $\{T_E^{(i)}\}^{4}_{i=1}$.  The \textbf{EXP} score is then computed as follows:
\begin{gather}
S^{*}_{S} = F(Q_{S}, \mathcal{G}^*(c), \mathcal{G}^*(e), \{T_S^{(i)}\}^{4}_{i=1})\\
f_S(S^{*}_{S}) = \begin{cases}
1, & S^{*}_{S} = {S^{*}_{S}}'\\
0, & \text{otherwise}
\end{cases}
\end{gather}
 where $S^{*}_{S}$ represents the MLLMs' prediction. $F$ represents MLLM. $f_S$ represents the function to calculate the \textbf{EXP} score. ${S^{*}_{S}}'$ represents the correct answer.

\section{Experiments}

\begin{figure*}[!t]
  \centering
  \vspace{-0.8cm}   \includegraphics[width=0.95\linewidth]{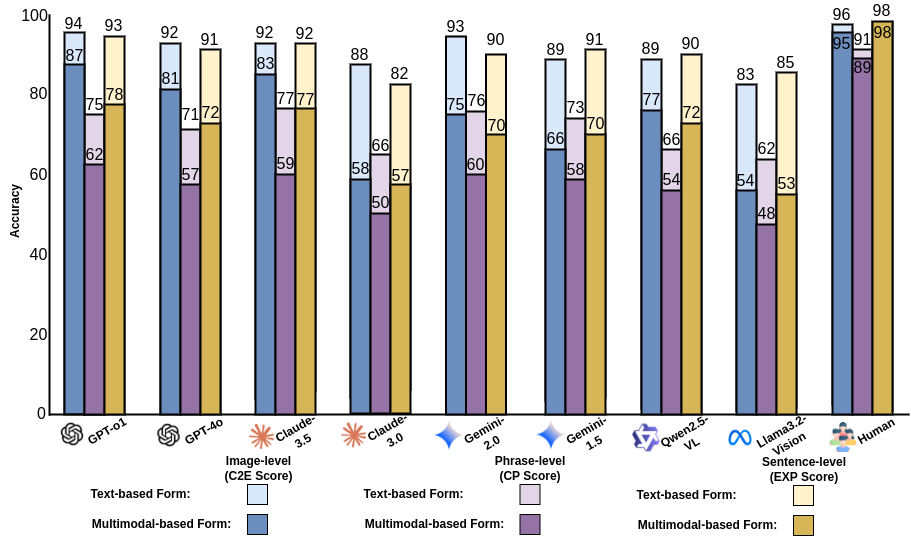}
   \vspace{-0.1cm}
   \caption{Main experimental results of several popular MLLMs on our MuCR benchmark. ``Human'' performance is represented by the average accuracy of ten attempts by volunteers.}
   \label{fig5}
   \vspace{-0.1cm}
\end{figure*}

\subsection{Experimental Setup}
\label{M4.1}
We evaluated several popular MLLMs on our MuCR benchmark, including GPT-o1~\cite{gpto1}, GPT-4o~\cite{gpt4o}, Claude-3.5~\cite{claude3.5}, Claude-3.0~\cite{claude3}, Gemini-2.0~\cite{gemini2.0}, Gemini-1.5~\cite{gemini1.5}, Qwen2.5-VL~\cite{yang2024qwen2}, and Llama3.2-Vision~\cite{llama3.2}. For the currently popular models, DeepSeek-R1~\cite{guo2025deepseek} and DeepSeek-V3~\cite{liu2024deepseek}, we did not fully evaluate their performance since their image readers currently only support extracting text from images without additional functionality (see Appendix~\ref{A2.1} for a comparison of their text-based performance). Additionally, we also considered some lightweight open-source models, including
LLaVA-NeXT~\cite{li2024llava}, OpenFlamingo-v2~\cite{awadalla2023openflamingo}, LLaVA-v1.6~\cite{liu2024visual}, MiniGPT4-v2~\cite{zhu2023minigpt}, and InstructBLIP~\cite{dai2023instructblip}. Since some models only accept a single image input, we provided all of them with a composite image composed of multiple smaller images, as shown in Figure~\ref{fig9} (a). Finally, we established a human performance baseline on the MuCR benchmark using crowd workers for comparison.

\subsection{Experimental Results}
Figure~\ref{fig5} presents the main results of popular MLLMs and human performance on the MuCR benchmark, leading to the following observations:
(1) \textbf{All models on MuCR lag behind human performance in both settings.} Among these models, GPT-o1~\cite{gpto1} achieves the highest scores, with 94\% on C2E score, 75\% on CP score, and 93\% on EXP score in the text condition, while 87\% on C2E, 62\% on CP, and 78\% on EXP in the multimodal condition. Nevertheless, these results still fall short of human performance, suggesting substantial room for improvement. (2) \textbf{All models exhibit a significant cross-modal performance gap}. All models show a noticeable drop in performance when handling multimodal causal inference, whereas humans do not. This discrepancy indicates potential factors restricting cross-modal generalization in MLLMs, likely stemming from the visual component, given that these models already demonstrate robust causal reasoning in text-based cases.

\begin{figure}[t]
  \centering
  \vspace{-0.2cm}   \includegraphics[width=0.95\linewidth]{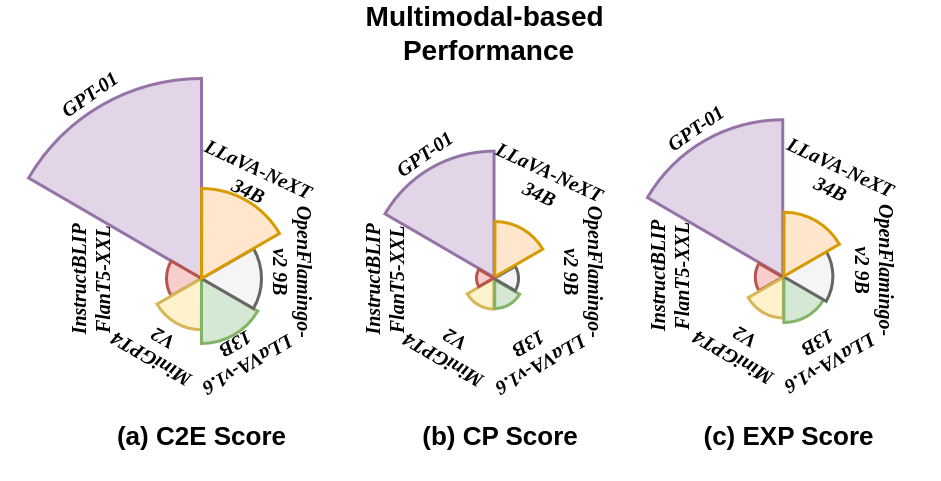}
   \vspace{-0.3cm}
   \caption{Experimental results of lightweight open-source models on the multimodal-based form. For detailed numbers see Table~\ref{Sheet3}. Best viewed by zooming in.}
   \label{fig6}
   \vspace{-0.3cm}
\end{figure}

Figure~\ref{fig6} presents the multimodal performance of various lightweight open-source models, revealing that they lag significantly behind GPT-o1. Among these, LLaVA-Next achieves the best results, with 29\% on C2E, 17\% on CP, and 21\% on EXP, which are only around the random selection baseline of 25\%. Compared to models like Llama3.2-Vision and Qwen2.5-VL, there is still considerable room for improvement for these models.

\section{Cross-modal Generalization Analysis and Enhancement}
In this section, we examine the factors that may affect cross-modal generalization. Building on previous findings that attribute these gaps primarily to the visual component, we focus on two main categories: visual format factors and visual semantic factors.
\begin{itemize}
    \item \textbf{Visual Format Factors.} These involve cases that share the same underlying semantics but differ in how they are visually presented, such as variations in picture style or the form of the visual input.
    \item \textbf{Visual Semantic Factors.} These involve cases with consistent visual formats but slight semantic differences, such as contextual variations in image details or the inclusion of additional text hints, resulting in richer semantic content.
\end{itemize}
In addition to investigating these cross-modal generalization factors, we also explore potential enhancement strategies based on our findings.

\subsection{Visual Format Factors}
\begin{figure}[!ht]
  \centering
  \vspace{-0.0cm}   \includegraphics[width=0.94\linewidth]{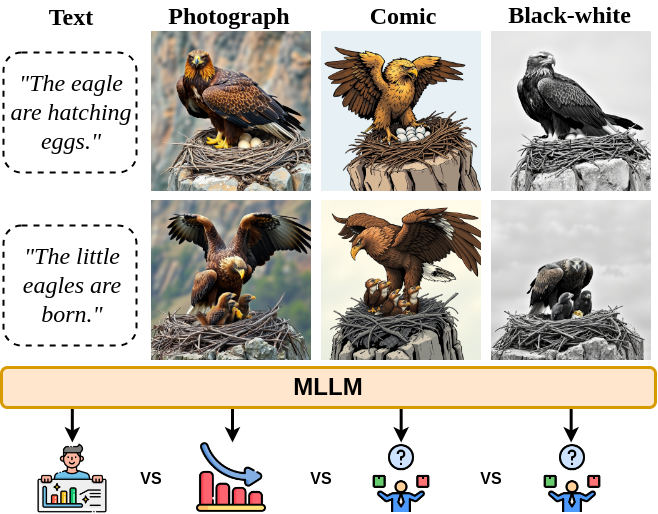}
   \vspace{-0.1cm}
   \caption{An example of cause and effect showing in three picture styles with the same semantic meanings.}
   \label{fig7}
   \vspace{-0.1cm}
\end{figure}

\paragraph{Picture Style.}
We investigate how different picture styles may affect causal reasoning. Figure~\ref{fig7}  

\begin{figure}[!ht]
  \centering
  \vspace{-0.4cm}   \includegraphics[width=0.94\linewidth]{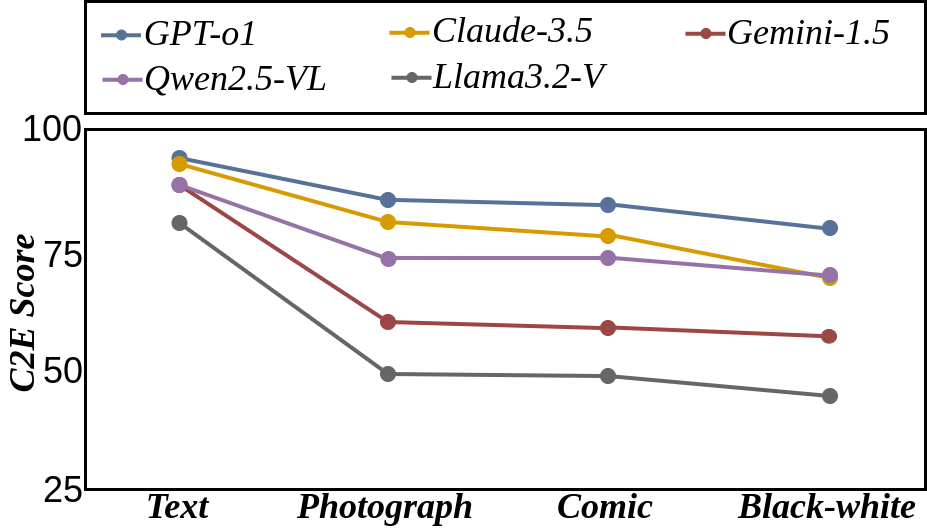}
   \vspace{-0.1cm}
   \caption{The C2E score of different models tested on three different picture styles.}
   \label{fig8}
   \vspace{-0.4cm}
\end{figure}

\begin{figure}[t]
  \centering
  \vspace{-0.3cm}   \includegraphics[width=0.95\linewidth]{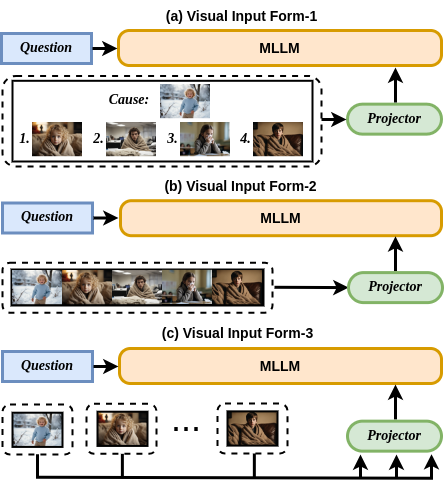}
   \vspace{-0.0cm}
   \caption{The illustration of three different visual input forms we examined.}
   \label{fig9}
   \vspace{-0.1cm}
\end{figure}

\begin{table}[t]
\centering
\renewcommand{\arraystretch}{0.5}
\resizebox{1.0\linewidth}{!}{
\begin{NiceTabular}{lcccc}[code-before=\rowcolor{rowgray}{2,6,10},cell-space-limits=1.1pt]
\specialrule{0.8pt}{0pt}{0.2pt}
\fontsize{5.5}{10}\selectfont Visual Input &\fontsize{5.5}{10}\selectfont Style   &\fontsize{5.5}{10}\selectfont C2E   &\fontsize{5.5}{10}\selectfont CP   &\fontsize{5.5}{10}\selectfont EXP  \\ \hline
\multicolumn{5}{c}{\fontsize{5.5}{10}\selectfont GPT-o1~\cite{gpto1}}              \\ \hline
\fontsize{5.5}{10}\selectfont Form-1       &\fontsize{5.5}{10}\selectfont  &\fontsize{5.5}{10}\selectfont 87.50 &\fontsize{5.5}{10}\selectfont 62.00 &\fontsize{5.5}{10}\selectfont 78.00 \\
\fontsize{5.5}{10}\selectfont Form-2       &\fontsize{5.5}{10}\selectfont Mixture &\fontsize{5.5}{10}\selectfont 84.25 &\fontsize{5.5}{10}\selectfont 60.50 &\fontsize{5.5}{10}\selectfont 79.00 \\
\fontsize{5.5}{10}\selectfont Form-3       &\fontsize{5.5}{10}\selectfont &\fontsize{5.5}{10}\selectfont 89.00 &\fontsize{5.5}{10}\selectfont 67.50 &\fontsize{5.5}{10}\selectfont 86.25 \\ 
\hline
\multicolumn{5}{c}{\fontsize{5.5}{10}\selectfont Claude-3.5~\cite{claude3.5}}         \\ \hline
\fontsize{5.5}{10}\selectfont Form-1       &\fontsize{5.5}{10}\selectfont  &\fontsize{5.5}{10}\selectfont 83.50 &\fontsize{5.5}{10}\selectfont 59.75 &\fontsize{5.5}{10}\selectfont 77.50 \\
\fontsize{5.5}{10}\selectfont Form-2       &\fontsize{5.5}{10}\selectfont Mixture &\fontsize{5.5}{10}\selectfont 53.50 &\fontsize{5.5}{10}\selectfont 36.00 &\fontsize{5.5}{10}\selectfont 68.50 \\
\fontsize{5.5}{10}\selectfont Form-3       &\fontsize{5.5}{10}\selectfont  &\fontsize{5.5}{10}\selectfont 85.00 &\fontsize{5.5}{10}\selectfont 66.75 &\fontsize{5.5}{10}\selectfont 82.25 \\ \hline
\multicolumn{5}{c}{\fontsize{5.5}{10}\selectfont Gemini-1.5~\cite{gemini1.5}}         \\ \hline
\fontsize{5.5}{10}\selectfont Form-1       &\fontsize{5.5}{10}\selectfont  &\fontsize{5.5}{10}\selectfont 66.50 &\fontsize{5.5}{10}\selectfont 58.50 &\fontsize{5.5}{10}\selectfont 70.50 \\
\fontsize{5.5}{10}\selectfont Form-2       &\fontsize{5.5}{10}\selectfont Mixture &\fontsize{5.5}{10}\selectfont 69.50 &\fontsize{5.5}{10}\selectfont 57.25 &\fontsize{5.5}{10}\selectfont 63.00 \\
\fontsize{5.5}{10}\selectfont Form-3       &\fontsize{5.5}{10}\selectfont  &\fontsize{5.5}{10}\selectfont 83.50 &\fontsize{5.5}{10}\selectfont 65.25 &\fontsize{5.5}{10}\selectfont 84.00 \\
\specialrule{0.8pt}{0pt}{0.2pt}
\end{NiceTabular}}
\vspace{-0.0cm}
\caption{The performance of different visual input forms on our MuCR benchmark. The mixture means we test on mixture picture style.}
\vspace{-0.3cm}
   \label{tab1}
\end{table}

\noindent shows an example of the same cause-and-effect scenario presented in three styles. As indicated by the results in Figure~\ref{fig8}, MLLMs perform similarly when presented with photographs and comic images, but with a slight drop for black-white images. Overall, altering the picture style while keeping the same semantic content has only a minimal effect on MLLMs' performance (see Appendix~\ref{A3.1} for detailed comparison).

\paragraph{Form of Visual Input.}
We also explore whether the structure of visual inputs affects the final output. Figure~\ref{fig9} illustrates the three types of visual input forms we examined. Table~\ref{tab1} presents the performance of three models on MuCR using these different formats. It indicates that all models get marked performance improvements. Our case analysis suggests that, compared to Form-3, Forms-1 and Form-2 restrict MLLMs’ ability to perceive certain details that could serve as crucial visual cues for enhancing multimodal causal reasoning (see Appendix~\ref{A3.2} for case studies).

\begin{figure}[t]
  \centering
  \vspace{-0.3cm}   \includegraphics[width=1.0\linewidth]{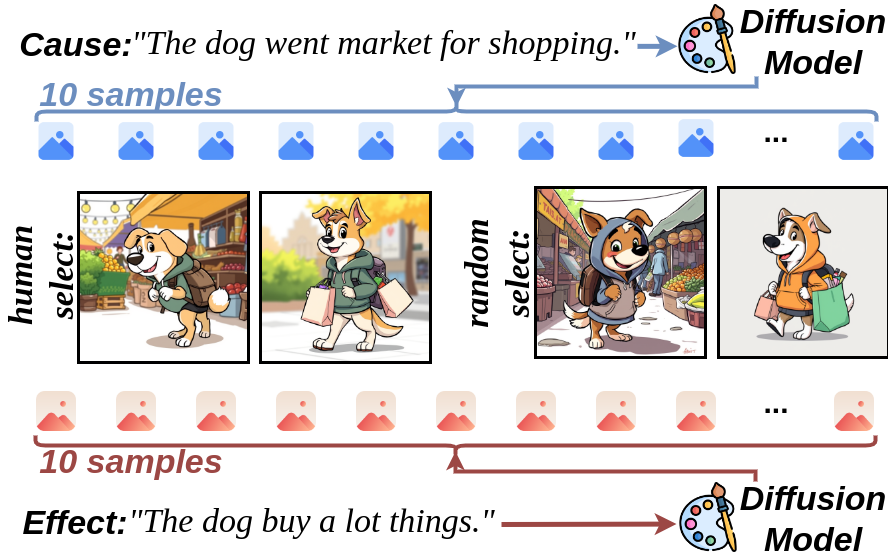}
   \vspace{-0.3cm}
   \caption{Two image pairs illustrate the same cause-and-effect relationship but exhibit different contextual correlations.}
   \label{fig10}
   \vspace{-0.2cm}
\end{figure}

\begin{figure}[t]
  \centering
  \vspace{-0.0cm}   \includegraphics[width=1.0\linewidth]{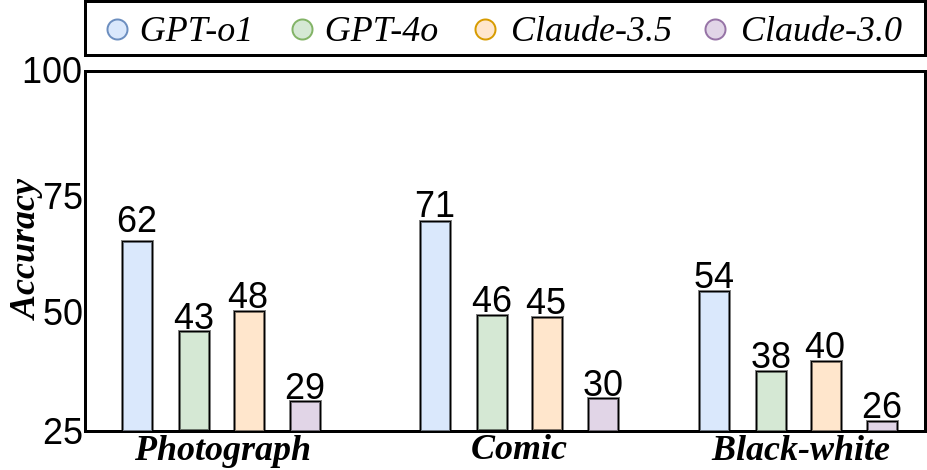}
   \vspace{-0.3cm}
   \caption{Using human selection as the standard, the models exhibit varying levels of selection accuracy.}
   \label{fig11}
   \vspace{-0.2cm}
\end{figure}

\subsection{Visual Semantic Factors}
\paragraph{Contextual Variation.}
In addition to examining visual format factors, we also explore whether visual semantics influence MLLMs' final output. As shown in Figure~\ref{fig1}, MLLMs, particularly GPT-o1, can identify visual cues such as action, appearance, and environment, and integrate these details into their causal inference process. Additionally, the case study in the above paragraph also confirms that visual cues are essential for accurate multimodal causal inference. To further investigate, we assess whether the ability to identify visual cues correlates with multimodal causal reasoning performance. For this purpose, we use manually selected siamese image pairs that best capture semantic causality and maintain visual consistency, along with some pairs that exhibit minor contextual variations (see Figure~\ref{fig10}). Our challenge is as follows: given a human-selected cause image, the models must identify the corresponding effect image from random 3 samples and 1 correct one. Figure~\ref{fig11} shows that among the four models tested, GPT-o1 excels at identifying visual cues, while Claude-3.0 performs the worst, with GPT-4.0 and Claude-3.5 falling in between (see Appendix~\ref{A3.3} for case studies). This finding confirms a positive correlation between an MLLM's ability to identify visual cues, distinguish contextual variations, and its overall multimodal causal reasoning performance.

\paragraph{Text Hints.}
Since we verified a positive correlation between multimodal causal reasoning and visual cue perception, the next question is whether text hints can compensate for shortcomings in visual cue perception. To explore this, we use the contextual descriptions generated during dataset creation as dense captions, as they provide detailed raw information while preserving correct semantic meanings. Table~\ref{tab2} shows that adding text hints significantly improves MLLMs' performance, suggesting that enhancing visual cue identification is a promising avenue for improving cross-modal generalization.

\begin{figure}[t]
  \centering
  \vspace{-0.0cm}   \includegraphics[width=0.95\linewidth]{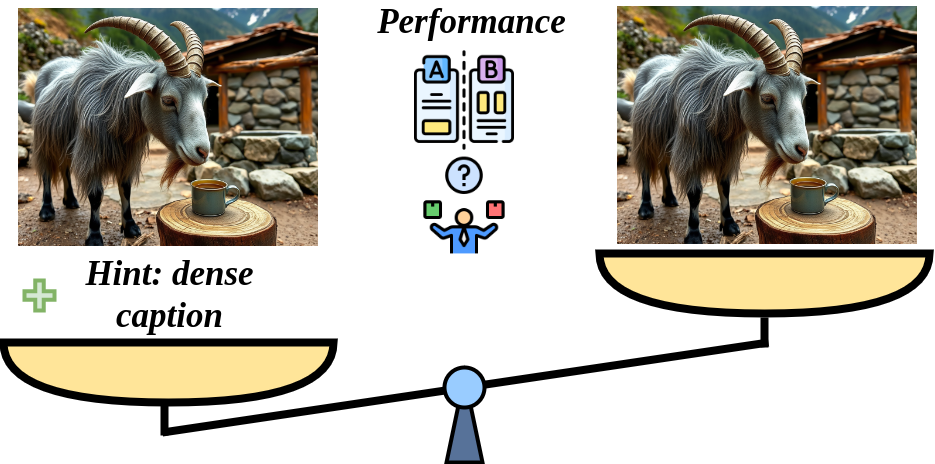}
   \vspace{-0.0cm}
   \caption{Does adding text hints compensate for shortcomings in visual cue perception?}
   \label{fig12}
   \vspace{-0.0cm}
\end{figure}

\begin{table}[t]
\centering
\renewcommand{\arraystretch}{0.5}
\resizebox{1.0\linewidth}{!}{
\begin{NiceTabular}{lcccc}[code-before=\rowcolor{rowgray}{2,5,8,11},cell-space-limits=1.1pt]
\specialrule{0.8pt}{0pt}{0.2pt}
\fontsize{5.5}{10}\selectfont Add Hints &\fontsize{5.5}{10}\selectfont Style   &\fontsize{5.5}{10}\selectfont C2E   &\fontsize{5.5}{10}\selectfont CP   &\fontsize{5.5}{10}\selectfont EXP  \\ \hline
\multicolumn{5}{c}{\fontsize{5.5}{10}\selectfont GPT-o1~\cite{gpto1}}              \\ \hline
\fontsize{5.5}{10}\selectfont Before       &\fontsize{5.5}{10}\selectfont  \multirow{2}{*}{Mixture} &\fontsize{5.5}{10}\selectfont 87.50 &\fontsize{5.5}{10}\selectfont 62.00 &\fontsize{5.5}{10}\selectfont 78.00 \\
\fontsize{5.5}{10}\selectfont After       &\fontsize{5.5}{10}\selectfont  &\fontsize{5.5}{10}\selectfont 91.25 &\fontsize{5.5}{10}\selectfont 69.50 &\fontsize{5.5}{10}\selectfont 88.50 \\
\hline
\multicolumn{5}{c}{\fontsize{5.5}{10}\selectfont GPT-4o~\cite{gpt4o}}              \\ \hline
\fontsize{5.5}{10}\selectfont Before       &\fontsize{5.5}{10}\selectfont  \multirow{2}{*}{Mixture} &\fontsize{5.5}{10}\selectfont 81.25 &\fontsize{5.5}{10}\selectfont 57.25 &\fontsize{5.5}{10}\selectfont 72.50 \\
\fontsize{5.5}{10}\selectfont After       &\fontsize{5.5}{10}\selectfont  &\fontsize{5.5}{10}\selectfont 89.00 &\fontsize{5.5}{10}\selectfont 66.50 &\fontsize{5.5}{10}\selectfont 87.50 \\
\hline
\multicolumn{5}{c}{\fontsize{5.5}{10}\selectfont Claude-3.5~\cite{claude3.5}}         \\ \hline
\fontsize{5.5}{10}\selectfont Before       &\fontsize{5.5}{10}\selectfont \multirow{2}{*}{Mixture} &\fontsize{5.5}{10}\selectfont 83.50 &\fontsize{5.5}{10}\selectfont 59.75 &\fontsize{5.5}{10}\selectfont 77.50 \\
\fontsize{5.5}{10}\selectfont After       &\fontsize{5.5}{10}\selectfont  &\fontsize{5.5}{10}\selectfont 87.50 &\fontsize{5.5}{10}\selectfont 68.50 &\fontsize{5.5}{10}\selectfont 86.00 \\ \hline
\multicolumn{5}{c}{\fontsize{5.5}{10}\selectfont Claude-3.0~\cite{claude3}}         \\ \hline
\fontsize{5.5}{10}\selectfont Before       &\fontsize{5.5}{10}\selectfont \multirow{2}{*}{Mixture} &\fontsize{5.5}{10}\selectfont 58.00 &\fontsize{5.5}{10}\selectfont 50.25 &\fontsize{5.5}{10}\selectfont 57.00 \\
\fontsize{5.5}{10}\selectfont After       &\fontsize{5.5}{10}\selectfont  &\fontsize{5.5}{10}\selectfont 73.00 &\fontsize{5.5}{10}\selectfont 59.50 &\fontsize{5.5}{10}\selectfont 77.00 \\
\specialrule{0.8pt}{0pt}{0.2pt}
\end{NiceTabular}}
\vspace{-0.0cm}
\caption{The impact of adding text hints on different models.}
\vspace{-0.1cm}
   \label{tab2}
\end{table}

\subsection{Generalization Enhancement} 
\begin{figure}[t]
  \centering
  \vspace{-0.0cm}   \includegraphics[width=0.98\linewidth]{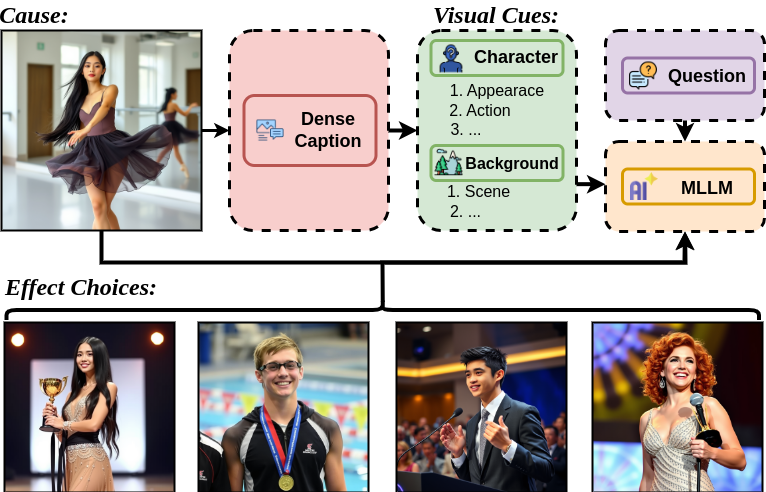}
   \vspace{-0.0cm}
   \caption{The structure of our VcCoT. Best viewed by zooming in.}
   \label{fig13}
   \vspace{-0.0cm}
\end{figure}

Based on our above analysis, the most crucial factor affecting MLLMs' cross-modal generalization is the ability to identify visual cues. In response, we propose VcCoT, a method designed to enhance visual cue identification for causal inference. Inspired by MMCoT~\cite{zhang2023multimodal} and CCoT~\cite{mitra2024compositional}, our approach first converts images into dense captions, then extracts visual details categorized as \emph{Character} and \emph{Background}. Finally, these cues guide the MLLMs' reasoning process, as illustrated in Figure~\ref{fig13}. Table~\ref{tab3} demonstrates that VcCoT achieves superior performance than others. We also show some qualitative results in Appendix~\ref{A3.5}.

\begin{table}[t]
\centering
\renewcommand{\arraystretch}{0.5}
\resizebox{1.0\linewidth}{!}{
\begin{NiceTabular}{lcccc}[code-before=\rowcolor{rowgray}{2,8},cell-space-limits=1.1pt]
\specialrule{0.8pt}{0pt}{0.2pt}
\fontsize{5.5}{10}\selectfont Strategy &\fontsize{5.5}{10}\selectfont Style   &\fontsize{5.5}{10}\selectfont C2E   &\fontsize{5.5}{10}\selectfont CP   &\fontsize{5.5}{10}\selectfont EXP  \\ \hline
\multicolumn{5}{c}{\fontsize{5.5}{10}\selectfont GPT-o1~\cite{gpto1}}              \\ \hline
\fontsize{5.5}{10}\selectfont Direct       &\fontsize{5.5}{10}\selectfont  \multirow{5}{*}{Mixture} &\fontsize{5.5}{10}\selectfont 87.50 &\fontsize{5.5}{10}\selectfont 62.00 &\fontsize{5.5}{10}\selectfont 78.00 \\
\fontsize{5.5}{10}\selectfont CoT       &\fontsize{5.5}{10}\selectfont  &\fontsize{5.5}{10}\selectfont 86.25 &\fontsize{5.5}{10}\selectfont 61.50 &\fontsize{5.5}{10}\selectfont 76.00 \\
\fontsize{5.5}{10}\selectfont CCoT       &\fontsize{5.5}{10}\selectfont  &\fontsize{5.5}{10}\selectfont 88.00 &\fontsize{5.5}{10}\selectfont 64.00 &\fontsize{5.5}{10}\selectfont 79.50 \\
\fontsize{5.5}{10}\selectfont MMCoT       &\fontsize{5.5}{10}\selectfont  &\fontsize{5.5}{10}\selectfont 84.25 &\fontsize{5.5}{10}\selectfont 60.5 &\fontsize{5.5}{10}\selectfont 86.50 \\
\fontsize{5.5}{10}\selectfont VcCoT       &\fontsize{5.5}{10}\selectfont  &\fontsize{5.5}{10}\selectfont 89.75 &\fontsize{5.5}{10}\selectfont 66.5 &\fontsize{5.5}{10}\selectfont 83.00 \\
\specialrule{0.8pt}{0pt}{0.2pt}
\end{NiceTabular}}
\vspace{-0.1cm}
\caption{The performance of different CoT strategies on MuCR benchmark. }
\vspace{-0.3cm}
   \label{tab3}
\end{table}

\section{Conclusion}
In this paper, we introduce MuCR, a novel multimodal causal reasoning benchmark that challenges MLLMs to discern causal links across different modalities by leveraging synthetic siamese images and text pairs. We also propose comprehensive metrics to assess MLLMs' understanding from multiple perspectives, including image-level alignment, phrase comprehension, and sentence-level explanation. Our experimental results reveal that current MLLMs exhibit a cross-modal gap in causal reasoning compared to their strong performance in purely textual settings. In-depth analysis highlights that effective visual cue identification is key to enhancing generalization, as MLLMs often struggle with implicit causal dependencies hidden in visual details. In response, we propose VcCoT, a method designed to improve visual cue identification for causal inference, with experimental results demonstrating its effectiveness.

\section{Limitation}
\begin{figure}[ht]
  \centering
  \vspace{-0.0cm}   \includegraphics[width=0.95\linewidth]{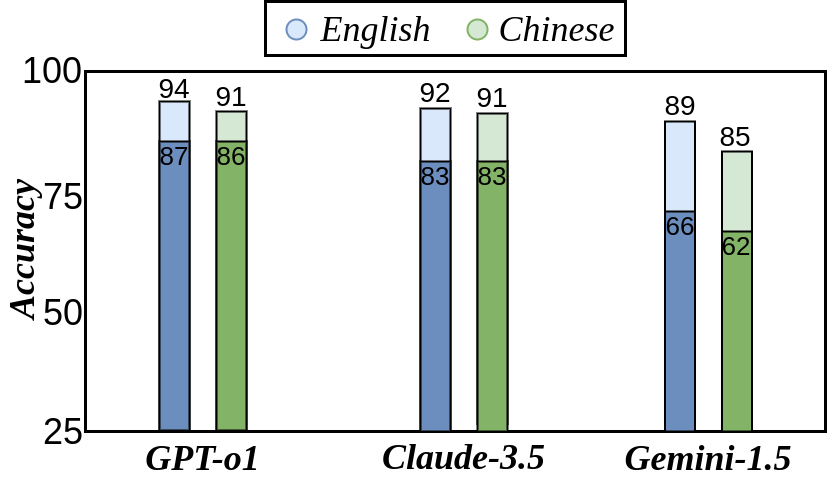}
   \vspace{-0.3cm}
   \caption{A comparison of different models on the C2E score with cross-linguistic setting.}
   \label{lim1}
   \vspace{-0.3cm}
\end{figure}

Although our research provides a comprehensive analysis of the potential factors affecting generalization from visual components, it has two notable limitations. First, as noted by \citet{wang2023cross}, cross-linguistic variations can influence performance and may require transfer learning. Figure~\ref{lim1} presents a simple comparison of transferring the question language from English to Chinese using the C2E score, indicating that cross-linguistic factors affect the final output of the models. However, due to human resource constraints, we did not extend this study to the CP and EXP scores, as these metrics require human reannotation of cue phrases and sentence explanations.

\begin{table}[ht]
\centering
\renewcommand{\arraystretch}{0.5}
\resizebox{1.0\linewidth}{!}{
\begin{NiceTabular}{lcccc}[code-before=\rowcolor{rowgray}{2,5,8,11},cell-space-limits=1.1pt]
\specialrule{0.8pt}{0pt}{0.2pt}
\fontsize{5.5}{10}\selectfont Fine-tune &\fontsize{5.5}{10}\selectfont Style   &\fontsize{5.5}{10}\selectfont C2E   &\fontsize{5.5}{10}\selectfont CP   &\fontsize{5.5}{10}\selectfont EXP  \\ \hline
\multicolumn{5}{c}{\fontsize{5.5}{10}\selectfont LLaVA-v1.6~\cite{liu2024visual}}              \\ \hline
\fontsize{5.5}{10}\selectfont Before       &\fontsize{5.5}{10}\selectfont  \multirow{2}{*}{Mixture} &\fontsize{5.5}{10}\selectfont 23.50 &\fontsize{5.5}{10}\selectfont 11.00 &\fontsize{5.5}{10}\selectfont 16.50 \\
\fontsize{5.5}{10}\selectfont After       &\fontsize{5.5}{10}\selectfont  &\fontsize{5.5}{10}\selectfont 20.00 &\fontsize{5.5}{10}\selectfont 13.75 &\fontsize{5.5}{10}\selectfont 15.25 \\
\hline
\multicolumn{5}{c}{\fontsize{5.5}{10}\selectfont MiniGPT4-v2~\cite{zhu2023minigpt}}              \\ \hline
\fontsize{5.5}{10}\selectfont Before       &\fontsize{5.5}{10}\selectfont  \multirow{2}{*}{Mixture} &\fontsize{5.5}{10}\selectfont 17.75 &\fontsize{5.5}{10}\selectfont 11.50 &\fontsize{5.5}{10}\selectfont 15.25 \\
\fontsize{5.5}{10}\selectfont After       &\fontsize{5.5}{10}\selectfont  &\fontsize{5.5}{10}\selectfont 19.00 &\fontsize{5.5}{10}\selectfont 13.50 &\fontsize{5.5}{10}\selectfont 16.00 \\
\hline
\multicolumn{5}{c}{\fontsize{5.5}{10}\selectfont InstructBLIP~\cite{dai2023instructblip}}         \\ \hline
\fontsize{5.5}{10}\selectfont Before       &\fontsize{5.5}{10}\selectfont \multirow{2}{*}{Mixture} &\fontsize{5.5}{10}\selectfont 12.25 &\fontsize{5.5}{10}\selectfont 6.50 &\fontsize{5.5}{10}\selectfont 9.50 \\
\fontsize{5.5}{10}\selectfont After       &\fontsize{5.5}{10}\selectfont  &\fontsize{5.5}{10}\selectfont 7.50 &\fontsize{5.5}{10}\selectfont 3.25 &\fontsize{5.5}{10}\selectfont 4.75 \\ 
\specialrule{0.8pt}{0pt}{0.2pt}
\end{NiceTabular}}
\vspace{-0.1cm}
\caption{The impact of direct fine-tuning on different models.}
\vspace{-0.3cm}
   \label{lim2}
\end{table}

Additionally, we explored fine-tuning a few lightweight open-source models. As shown in Table~\ref{lim2}, direct fine-tuning with the correct choices did not improve and in some cases even decreased the performance of these models. Our observations indicate that these models fail to capture the causal links between cause-and-effect images through fine-tuning. Notably, InstructBLIP even lost its ability to caption images accurately, exhibiting severe hallucinations. Due to limited resources, we did not investigate whether reinforcement learning~\cite{guo2025deepseek} or alternative strategies~\cite{Niklas2025s1} could further address the generalization problem on larger models such as Qwen2.5-VL~\cite{yang2024qwen2} or LLama3.2-Vision~\cite{llama3.2}.

\bibliography{acl}

\clearpage
\appendix
\large{\textbf{Appendix}}
\normalsize
\section{The MuCR Dataset}
\subsection{Task Formulation}
 As shown in Figure~\ref{fig2} (b), our dataset $\mathcal{D}:=\{(Q, \mathcal{G}^*(A), \{B^{(i)}\}_{i=1}^4)^{(k)}\}_{k=1}^N$ consisting of $N$ triples, each contains a question $Q$, an input $\mathcal{G}^*(A)$ (where $*$ represents the input form), and four potential choices $\{B^{(i)}\}_{i=1}^4$. The MLLMs are required to according to the question $Q$ and an input $\mathcal{G}^*(A)$ to select the correct answer from four potential choices $\{B^{(i)}\}_{i=1}^4$. The goal of this benchmark is to determine whether the input form ($*$) affects the MLLMs’ prediction accuracy. To this end, the biggest challenge is defined as follows:
 \begin{equation}
     \mathcal{G}^{text}(A) \approxtext{semantic}  \mathcal{G}^{multi}(A),
 \end{equation}
where $\approxtext{semantic}$ means $\mathcal{G}^*(A)$ retains identical or closely aligned semantic meaning across different modalities. To address this, we propose a novel transfer strategy that harnesses the linguistic capabilities of LLMs alongside the image generation abilities of diffusion models, effectively preserving semantic content while altering the input form.

\subsection{Overall Structure}
\label{A1.0}
\begin{figure}[t]
  \centering
  \vspace{-0.0cm}   \includegraphics[width=1.0\linewidth]{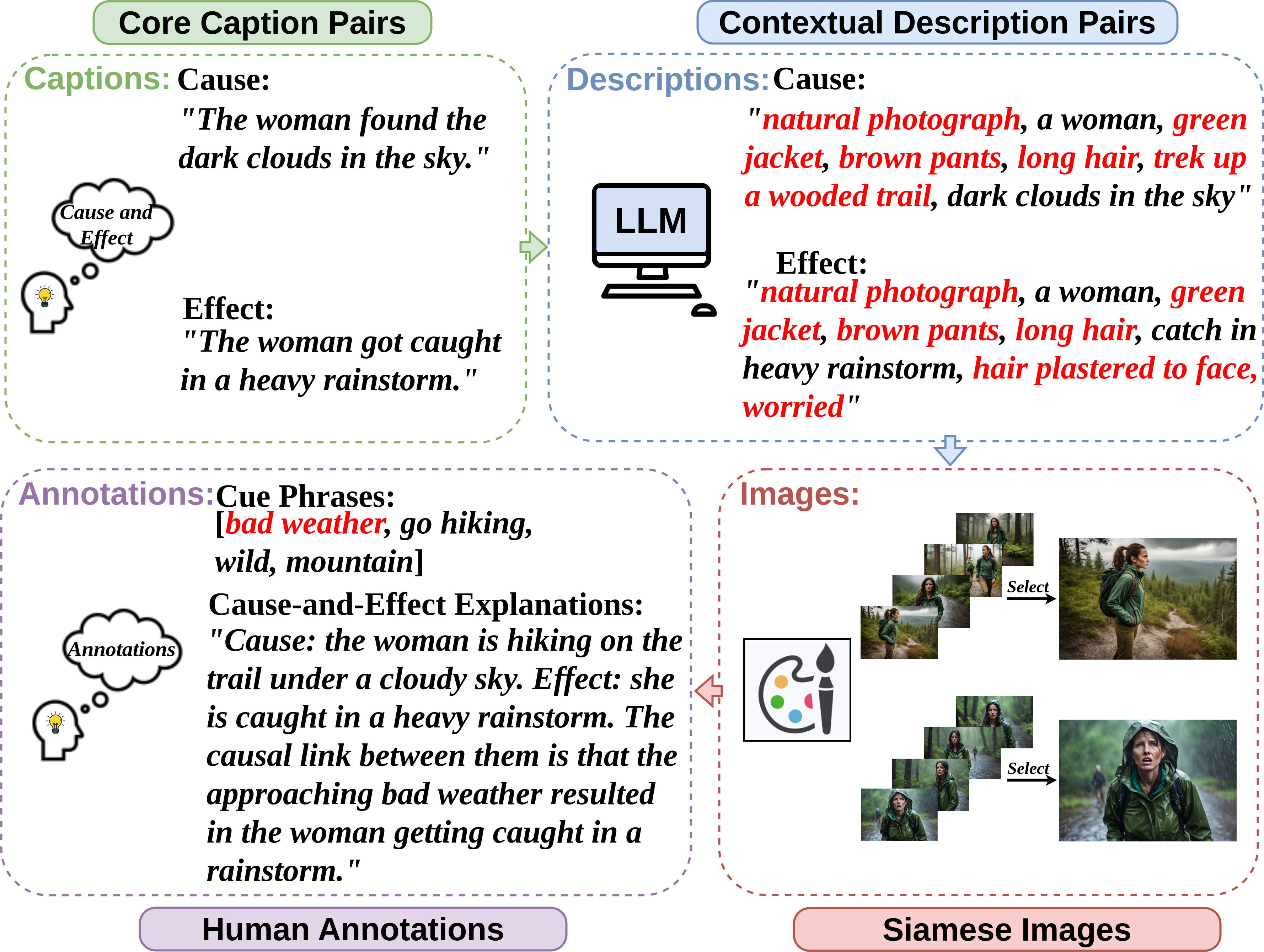}
   \vspace{-0.3cm}
   \caption{A detailed example of generating our MuCR dataset. Best viewed by zooming in.}
   \label{img1}
   \vspace{-0.1cm}
\end{figure}

Section~\ref{M3.0} only illustrates the simplified process of our MuCR benchmark generation. Here, we delve into more details about the generation process and the corresponding prompts. Figure~\ref{img1} showcases the detailed generation process of a weather-related causal case in our MuCR dataset. Our process begins with generating core caption pairs, each consisting of one caption describing the cause and the other stating the effect. We then leverage the language capabilities of LLMs to entail these paired captions into contextually relevant descriptions, enhancing the consistency of sentences to facilitate the creation of cause-and-effect image pairs. Then, we employ diffusion models to generate numerous Siamese images based on these descriptions. Finally, we annotate cue phrases and causality explanations for each pair.

\subsection{Generating Core Caption Pairs}
\label{A1.1}
\begin{figure}[t]
  \centering
  \vspace{-0.0cm}   \includegraphics[width=1.0\linewidth]{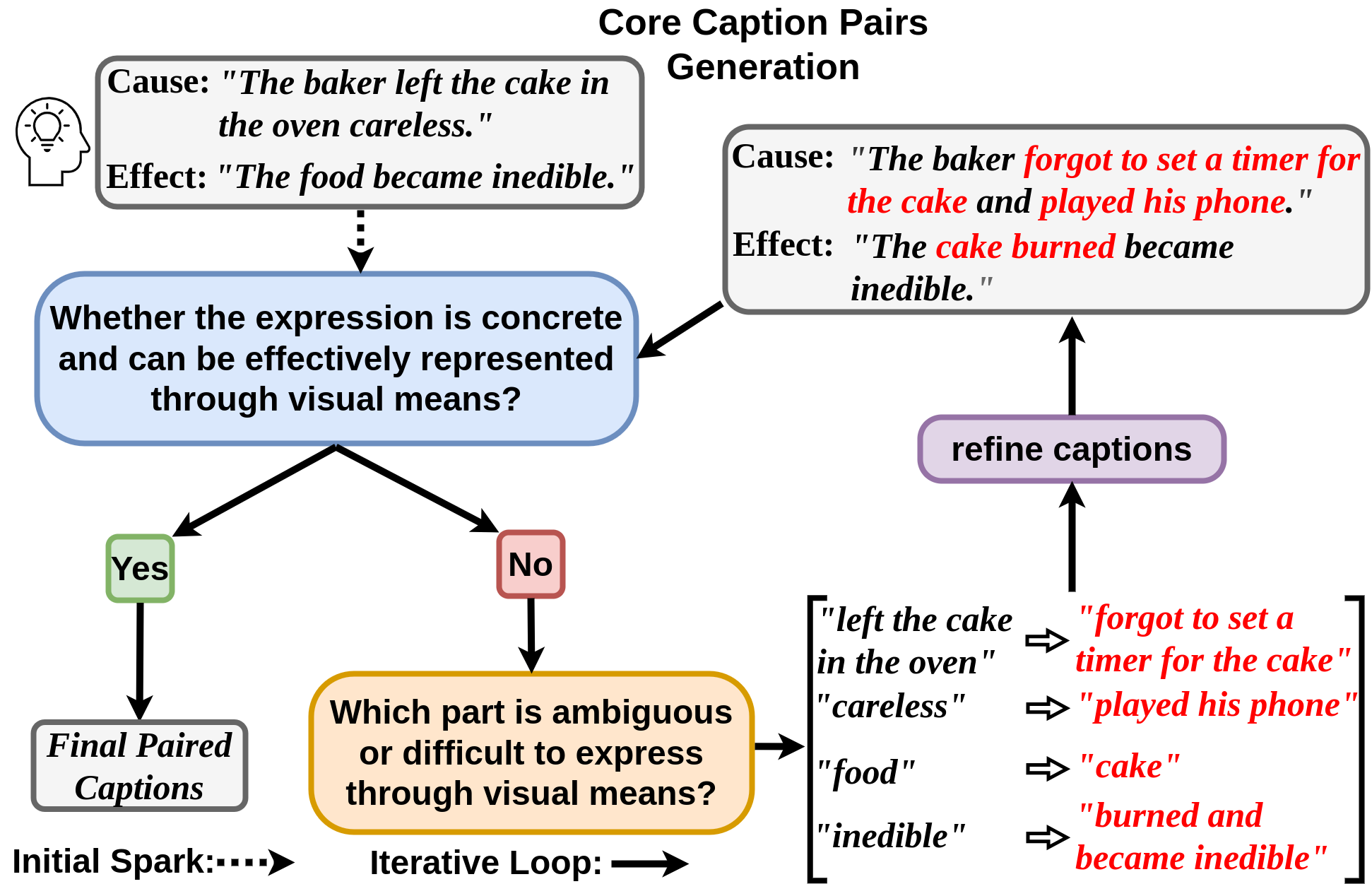}
   \vspace{-0.3cm}
   \caption{The process of generating paired captions through refinement loops, ensuring the final captions are precise and can be effectively represented through visual means.}
   \label{img2}
   \vspace{-0.1cm}
\end{figure}

Our MuCR benchmark begins with the creation of core caption pairs, where one caption outlines the cause and the other describes the effect. These pairs maintain semantic causality and serve two roles. First, they function as textual causal inference cases to challenge MLLMs’ textual reasoning ability. Second, they guide the subsequent synthesis of Siamese images. As shown in Figure~\ref{img2}, we employ a structured refinement loop that transforms initial brainstorming ideas into precise caption pairs, clearly depicting the cause-and-effect relationships. This process is guided by the principle: ``Whether the expression is concrete and can be effectively represented through visual means". Here, we discuss the rationale behind this rule and explain why volunteers are instructed to create core caption pairs in accordance with it. 

\begin{figure}[t]
  \centering
  \vspace{-0.0cm}   \includegraphics[width=1.0\linewidth]{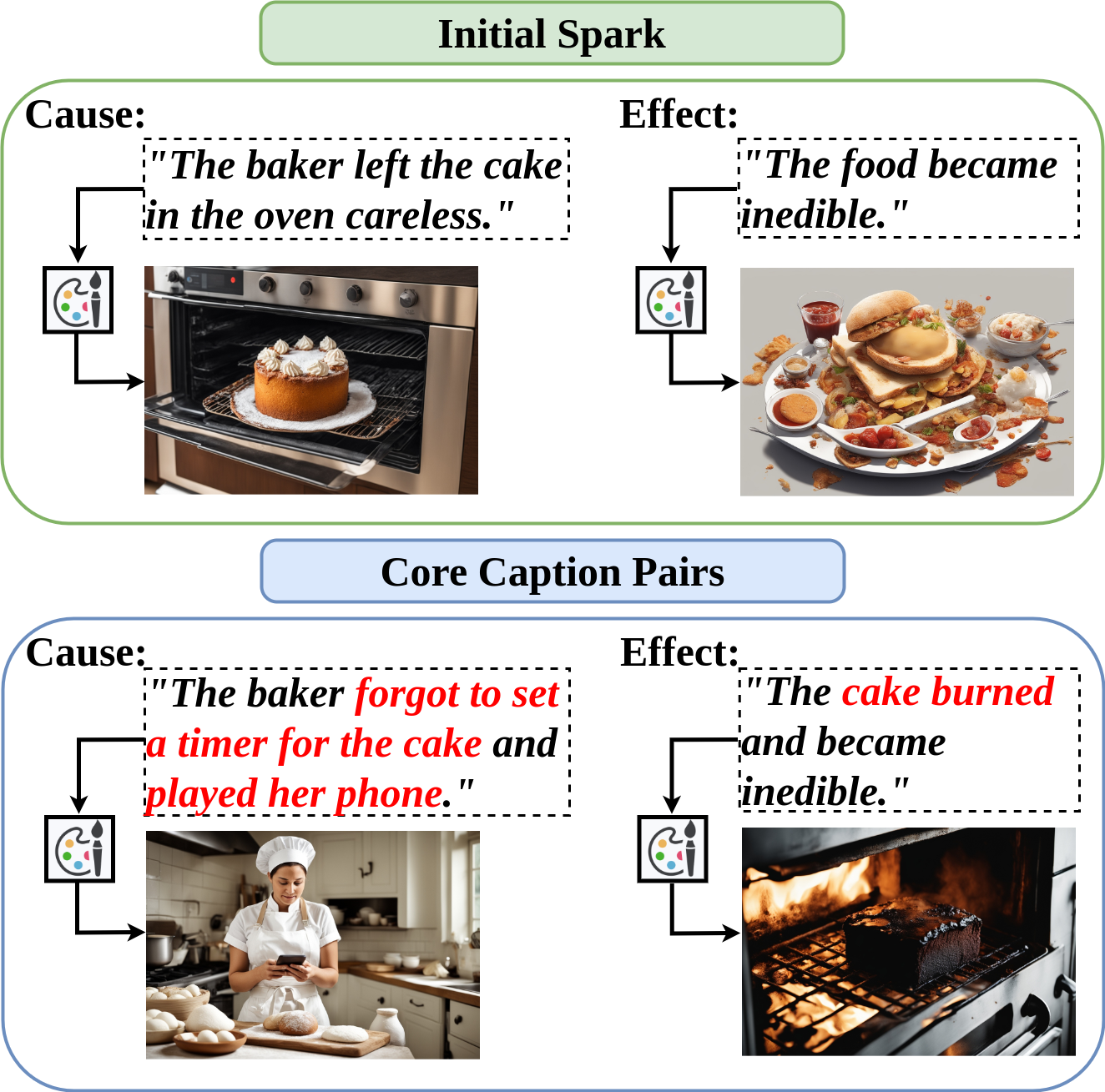}
   \vspace{-0.3cm}
   \caption{A comparison of directly using initial spark and our core caption pairs to generate cause-and-effect images through the diffusion model.}
   \label{img3}
   \vspace{-0.1cm}
\end{figure}

Figure~\ref{img3} compares the initial spark and core caption pairs in image synthesis. The comparison reveals that the initial spark often contains semantically ambiguous elements, leading to visual gaps in the generated images. For instance, the phrase ``the baker left the cake in the oven" might result in an image depicting only a cake in the oven, as the diffusion model may struggle to interpret or visually represent the action ``left". Another issue is subject conflict. For example, the phrase ``the food became inedible" might simply produce an image of unappealing food on a plate. However, within a cause-and-effect scenario, a human would easily infer that ``food" refers specifically to the ``cake." In contrast, our core caption pairs resolve these ambiguities by translating them into more concrete actions, such as replacing ``careless" with ``played his phone." This refinement significantly improves the quality of the generated images and the semantic causality between the pairs.

\begin{table*}[t]
    \centering
    \scalebox{0.8}{
    \begin{tabular}{llc}
     \hline
     \textbf{Cause} & \textbf{Effect} & \textbf{Category} \\
     \hline
     The man drove his car \textbf{at an excessive speed}. & The man \textbf{got a speeding ticket}. & Person \\
     The woman drove her car \textbf{at at an excessive speed}. & The woman was \textbf{pulled over by the policeman}. & Person \\
     The old man drave his car \textbf{at an excessive speed}. & The old man was \textbf{pulled over by the policeman}. & Person \\
     The old woman drave her car \textbf{at an excessive speed}. & The old woman \textbf{got a speeding ticket}. & Person \\
     \hline
     The shark is \textbf{attacking} the fish. & The fish \textbf{got wounded} on its back. & Animal \\
     The shark is \textbf{attacking} the seal. & The seal \textbf{got wounded} on its back. & Animal \\
     The shark is \textbf{attacking} the manta ray. & The manta ray \textbf{got wounded and bleeding}. & Animal \\
     The shark is \textbf{attacking} the penguin. & The penguin \textbf{got wounded and bleeding}. & Animal \\
     \hline
     The chrysanthemum is \textbf{blooming}. & The chrysanthemum \textbf{attracting bees to collect nectar}. & Plant \\
     The tulip is \textbf{blooming}. & The tulip \textbf{attracting bees to collect nectar}. & Plant \\
     The rose is \textbf{blooming}. & The rose \textbf{attracting bees to collect nectar}. & Plant \\
     The jasmine is \textbf{blooming}. & The jasmine \textbf{attracting bees to collect nectar}. & Plant \\
     \hline
     The rabbit \textbf{worked hard}. & The rabbit \textbf{earn much money}. & Character \\
     The monkey \textbf{worked hard}. & The monkey \textbf{earn much money}. & Character \\
     The bear \textbf{worked hard}. & The bear \textbf{earn much money}. & Character \\
     The fox \textbf{worked hard}. & The fox \textbf{earn much money}. & Character \\
     \hline
     The gardener \textbf{planted a tree}. & The tree \textbf{grew tall}. & Mixture \\
     The farmer \textbf{planted seeds}. & The seeds \textbf{sprouted into crops}. & Mixture \\
     The child \textbf{planted flowers}. & The flowers \textbf{bloomed in the garden}. & Mixture \\
     The woman \textbf{planted herbs}. & The herbs \textbf{grew in the pot}. & Mixture \\
     \hline
    \end{tabular}}
    \caption{Case studies for the paired caption generation process.}
    \label{sheet1}
\end{table*}

We ask the volunteers to design four paired captions as a group, each sharing similar causalities but containing different visual cues. These groups are intended to explore the capability of distinguishing similar causalities occurring in different subjects across various scenarios. Furthermore, to maintain the diversity of our dataset, we include a portion of non-human cases. While many causality scenarios feature humans as subjects, we also incorporate cases involving animals, plants, comic characters, and their interactions. Table~\ref{sheet1} shows generated paired-caption examples (i.e., four captions sharing similar causalities but involving different visual cues are paired as a group) for various scenarios (i.e., cases involving humans, animals, plants, comic characters, and mixtures).
Abstract expressions are concretized during the paired-caption generation process according to the causality. For instance, the scenario ``driving at excessive speed" is rephrased in terms of its potential outcomes, such as ``getting a speeding ticket" or ``being pulled over by a police officer". Similarly, the concept of ``blooming" is illustrated through its possible consequence, ``attracting bees to gather nectar". This process leverages causal reasoning to ground abstract ideas in real-world outcomes, thereby enhancing the intelligibility and reproducibility of the generated captions.

\subsection{Producing Contextual Description Pairs}
\label{A1.2}
\begin{figure}[t]
  \centering
  \vspace{-0.0cm}   \includegraphics[width=0.95\linewidth]{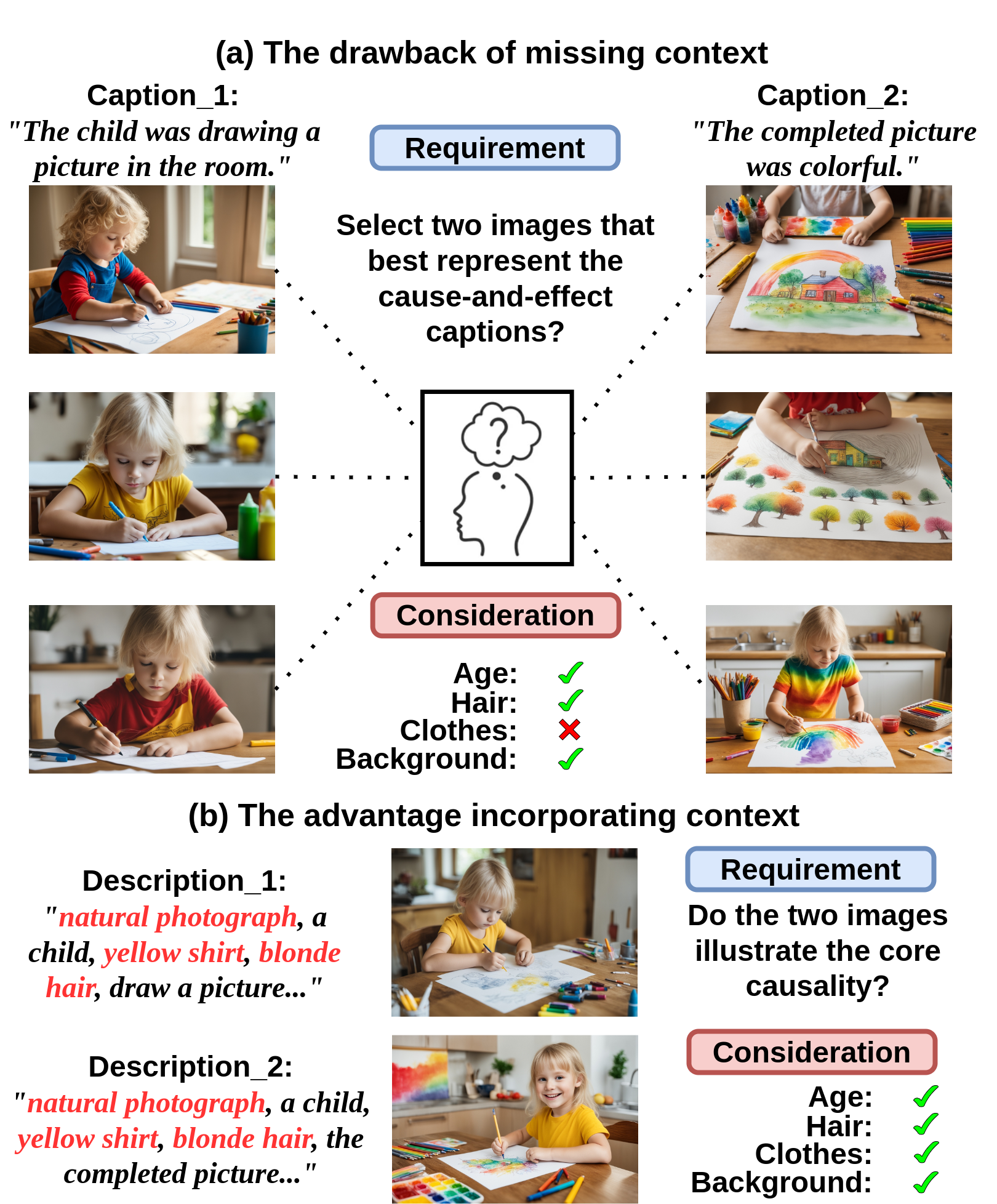}
   \vspace{-0.1cm}
   \caption{An example of core captions vs contextual descriptions in cause-and-effect image synthesis.}
   \label{img4}
   \vspace{-0.5cm}
\end{figure}

The absence of crucial visual cues could introduce randomness in image creation, which may lead to inconsistencies and potentially undermine the perceived causality between the siamese images. Recent research on content consistency~\cite{chen2025Ouroboros, Long2024VideoStudioGC} has become popular in long video generation by maintaining coherent content across frames. For image content consistency, Figure~\ref{img4} highlights the drawbacks of missing context and the advantages of incorporating context. As shown in Figure~\ref{img4} (a), although the two columns of images accurately represent the core caption, mismatched clothing disrupts the sense of causality, making it difficult to form coherent pairs. In contrast, the example in Figure~\ref{img4} (b) demonstrates that incorporating contextual information and transforming core captions into contextual descriptions effectively resolves this issue and reduces randomness in image synthesis. To achieve this, we leverage the linguistic capabilities of LLMs to enhance core caption pairs by enriching contextual details such as appearance, clothing, environment, and atmosphere. Additionally, we introduce subtle changes, such as variations in facial expressions, within the contextual description pairs to reflect the passage of time. These detailed variations emphasize the impact of causality over time, making the connection between siamese images more natural and coherent.

\begin{figure}[!t]
  \centering
  \vspace{-0.0cm}   \includegraphics[width=\linewidth]{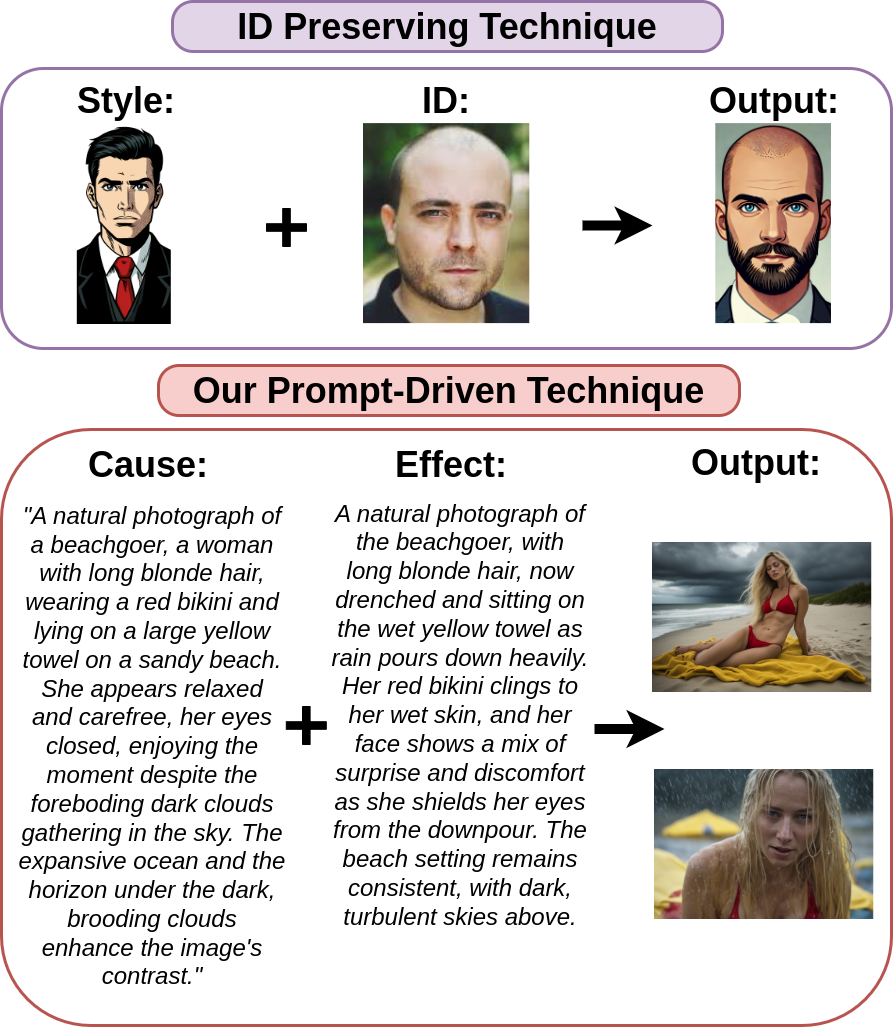}
   \vspace{-0.3cm}
   \caption{A comparison of identity-preserving technique and our prompt-driven technique on image synthesis.}
   \label{img5}
   \vspace{-0.5cm}
\end{figure}

We also compare identity-preserving techniques with our prompt-guidance method (Figure~\ref{img5}). Traditional identity-preserving image synthesis methods (e.g., LCM~\cite{gal2024lcm} and IP-Adapter~\cite{ye2023ip}) focus on image personalization by retaining identity details through a region encoder during the generation process~\cite{wang2024instantid}. However, this approach leads to two major issues. First, most existing identity-preserving techniques rely heavily on guided images, which limits their capacity for semantically-driven image generation and requires finding a suitable ID image for each causal scenario. Second, as the name suggests, identity-preserving methods focus primarily on maintaining facial identity (appearance) but struggle to incorporate cause-and-effect relationships across images. In contrast, our causal-and-effect image synthesis approach leverages the linguistic capabilities of large language models (LLMs) to integrate a richer spectrum of contextual information. 
It not only preserves human facial identity (appearance), but also accounts for additional details (e.g., clothing, environment, and overall atmosphere). This ensures that images remain coherent even when modifications are introduced through causal reasoning. The producing contextual description pairs prompt is organised as:  \textit{"Task Overview: You need to convert causal-relevant captions into detailed image descriptions for generating images. Ensure the following: (A) Consistency in Characters: Use one sentence to describe the person's appearance in each image description. Make sure two sentences in the two descriptions share the same information without using words like same or similar. (B) Face expression in Characters: If the descriptions contain the person's appearance, please directly add a sentence following to describe the person's facial expression, matching the scene. (C) Activities or Behaviors: Cause descriptions should exclusively detail the causal activities or behaviors, while effect descriptions should exclusively detail the resultant activities or behaviors. (D) Consistency in Scenes: If the scene remains unchanged between cause and effect, ensure the background description is identical in both descriptions. (E) Clear Causal Link: (1) Enhance Cause: Provide concrete details about what led to the effect. (2) Improve Effect: Ensure the effect is both visually and logically linked to the cause, using relatable or observable descriptions. "}

\subsection{Siamese Images and Annotations}
\label{A1.3}
In this section, we show some high-quality examples as follows:

\begin{figure*}
    \centering
    \includegraphics[width=0.95\textwidth]{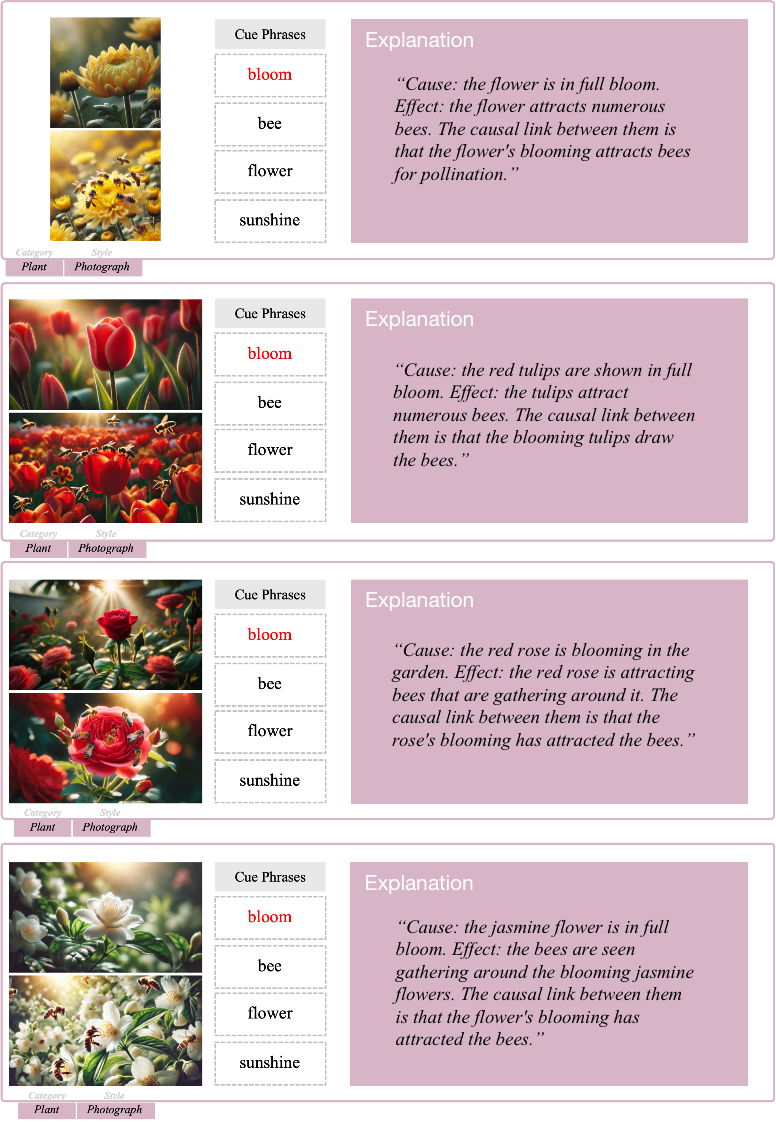}
    \caption{Example 1 - Plant}
    \label{plant}
\end{figure*}

\begin{figure*}
    \centering
    \includegraphics[width=0.95\textwidth]{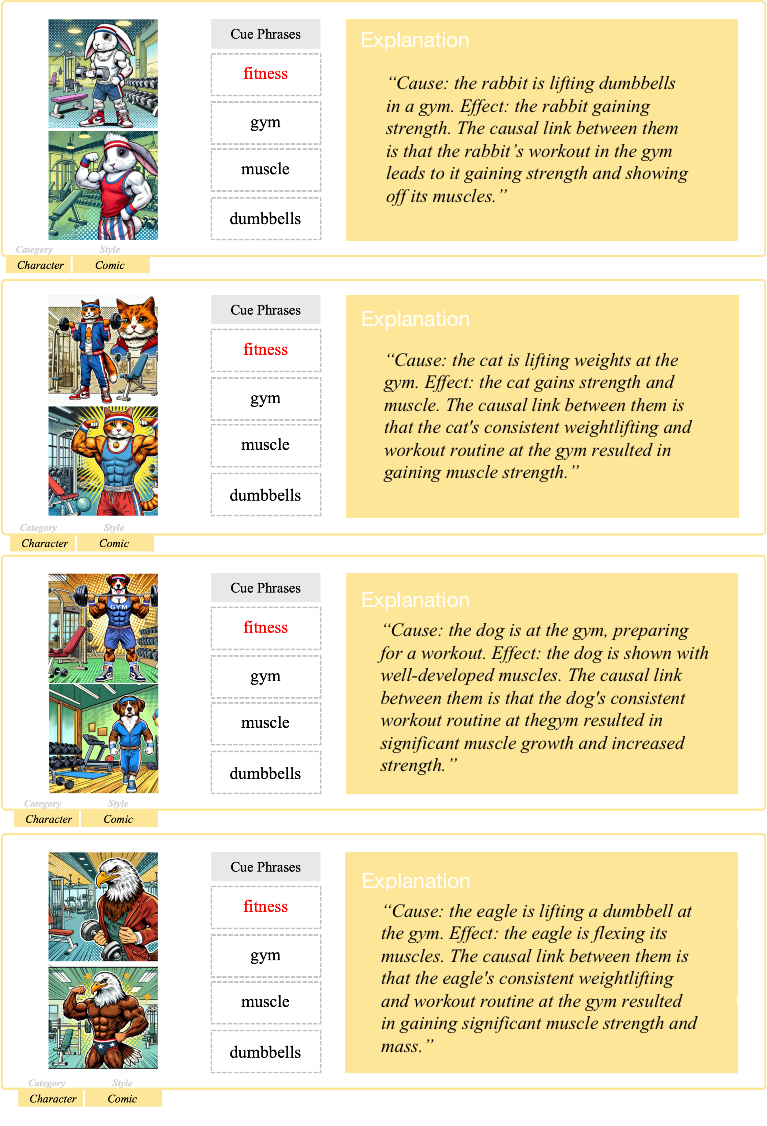}
    \caption{Example 2 - Character}
    \label{character}
\end{figure*}

\begin{figure*}
    \centering
    \includegraphics[width=0.95\textwidth]{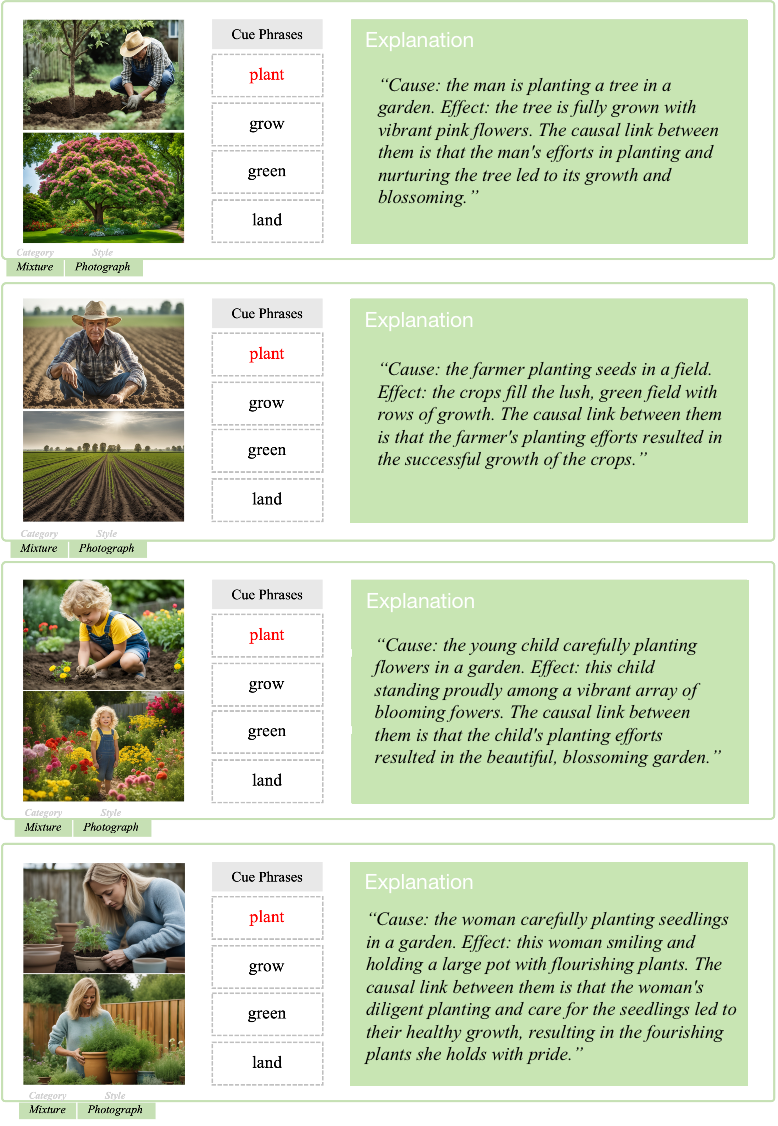}
    \caption{Example 3 - Mixture}
    \label{mixture}
\end{figure*}

\begin{table*}[t]
\centering
\scalebox{0.92}{
\begin{NiceTabular}{lcccccc}[code-before=\rowcolor{rowgray}{3,12},code-before=\rowcolor{darkgray}{18}, cell-space-limits=1.3pt]
\specialrule{1.2pt}{0pt}{0.2pt}
\multirow{2}{*}{Model} & \multicolumn{3}{c}{Text-based Form} & \multicolumn{3}{c}{Multimodal-based Form} \\ \cline{2-7} 
                       & C2E        & CP         & EXP       & C2E          & CP           & EXP         \\  \hline
\multicolumn{7}{c}{Popular MLLMs}                                                                        \\ \hline
GPT-o1~\cite{gpto1}                 & 94.00         & 75.50       & 93.00        & 87.50         & 62.00           & 78.00          \\
GPT-4o~\cite{gpt4o}                 & 92.75      & 71.75      & 91.50      & 81.25        & 57.25        & 72.50        \\
Claude-3.5~\cite{claude3.5}             & 92.50       & 77.00         & 92.75     & 83.50         & 59.75        & 77.5        \\
Claude-3.0~\cite{claude3}             & 88.25      & 66.75      & 82.00        & 58.00           & 50.25        & 57.00          \\
Gemini-2.0~\cite{gemini2.0}             & 93.00         & 76.00         & 90.50      & 75.50         & 60.75        & 70.25       \\
Gemini-1.5~\cite{gemini1.5}             & 89.00         & 73.25      & 91.50      & 66.50         & 58.50         & 70.75       \\
Qwen2.5-VL~\cite{yang2024qwen2}             & 89.00         & 66.00         & 90.00        & 77.00           & 54.50         & 72.00          \\
Llama3.2-Vision~\cite{llama3.2}        & 83.50       & 62.50       & 86.00        & 54.00           & 48.25        & 53.25       \\ \hline
\multicolumn{7}{c}{Lightweight Open-source Models}                                                       \\ \hline
LLaVA-NeXT~\cite{li2024llava}             & 54.50      & 37.50      & 48.00     & 29.00        & 17.00        & 21.00       \\
OpenFlamingo-v2~\cite{awadalla2023openflamingo}        & 23.00      & 16.00      & 17.25     & 20.00        & 9.75         & 18.00       \\
LLaVA-v1.6~\cite{liu2024visual}             & 25.25      & 17.25      & 18.00     & 23.50        & 11.00        & 16.50       \\
MiniGPT4-v2~\cite{zhu2023minigpt}            & 13.50      & 18.50      & 16.75     & 17.75        & 11.50        & 15.25       \\
InstructBLIP~\cite{dai2023instructblip}           & 14.50      & 10.00      & 8.50      & 12.25        & 6.50         & 9.50        \\ \hline
Human                  & 96.75      & 91.00      & 98.50     & 95.50        & 89.50        & 98.50      \\
\specialrule{1.2pt}{0pt}{0.2pt}
\end{NiceTabular}}
\caption{Main experimental results of different models on our MuCR benchmark.}
\label{Sheet3}
\end{table*}

In the plant category, as shown in Figure~\ref{plant}, take the jasmine flower pair: the cause image shows a blooming jasmine flower, while the effect image features a group of bees swarming around it. For this pair, we select ``bloom" as the positive cue phrase and ``bee", ``flower", and ``sunshine" as the negative ones, aligning with the visual information. The annotation emphasizes the connection between the flower's blooming and the attraction of bees.

In the character category, as shown in Figure~\ref{character}, consider the cat pair: the first image shows a cat lifting weights at the gym, while the second image depicts the cat gaining strength and muscle. For this, ``fitness" is used as the positive cue phrase, with ``gym", ``muscle", and ``dumbbells" as the negative ones, matching the visual content. The annotation focuses on the connection between consistent workouts and muscle gains.

In the mixture category, as shown in Figure~\ref{mixture}, take the female planting pair: the cause image shows a woman planting seedlings in a garden, while the effect image displays the same woman smiling and holding a large pot of flourishing plants. Here, ``plant" is the positive cue phrase, and ``grow", ``green", and ``land" are the negative ones, aligning with the visual information. The annotation emphasizes the relationship between her nurturing care and the plant's growth, along with her pride.

\begin{table}[t]
\centering
\renewcommand{\arraystretch}{0.5}
\resizebox{1.0\linewidth}{!}{
\begin{NiceTabular}{lcccc}[code-before=\rowcolor{rowgray}{2,5,7},cell-space-limits=1.1pt]
\specialrule{0.8pt}{0pt}{0.2pt}
\fontsize{5.5}{10}\selectfont Input Form &\fontsize{5.5}{10}\selectfont Style   &\fontsize{5.5}{10}\selectfont C2E   &\fontsize{5.5}{10}\selectfont CP   &\fontsize{5.5}{10}\selectfont EXP  \\ \hline
\multicolumn{5}{c}{\fontsize{5.5}{10}\selectfont GPT-o1~\cite{gpto1}}              \\ \hline
\fontsize{5.5}{10}\selectfont Text       &\fontsize{5.5}{10}\selectfont \textbackslash{}  &\fontsize{5.5}{10}\selectfont 94.00 &\fontsize{5.5}{10}\selectfont 75.50 &\fontsize{5.5}{10}\selectfont 93.00 \\
\fontsize{5.5}{10}\selectfont Image       &\fontsize{5.5}{10}\selectfont Mixture &\fontsize{5.5}{10}\selectfont 87.50 &\fontsize{5.5}{10}\selectfont 62.00 &\fontsize{5.5}{10}\selectfont 78.00 \\
\hline
\multicolumn{5}{c}{\fontsize{5.5}{10}\selectfont DeepSeek-R1~\cite{guo2025deepseek}}         \\ \hline
\fontsize{5.5}{10}\selectfont Text       &\fontsize{5.5}{10}\selectfont  \textbackslash{} &\fontsize{5.5}{10}\selectfont 96.00 &\fontsize{5.5}{10}\selectfont 73.50 &\fontsize{5.5}{10}\selectfont 95.00 \\ \hline
\multicolumn{5}{c}{\fontsize{5.5}{10}\selectfont DeepSeek-V3~\cite{li2014deepreid}}         \\ \hline
\fontsize{5.5}{10}\selectfont Text       &\fontsize{5.5}{10}\selectfont  \textbackslash{}&\fontsize{5.5}{10}\selectfont 91.50 &\fontsize{5.5}{10}\selectfont 72.25 &\fontsize{5.5}{10}\selectfont 92.00 \\
\specialrule{0.8pt}{0pt}{0.2pt}
\end{NiceTabular}}
\vspace{-0.1cm}
\caption{The performance comparison between GPT-o1 and DeepSeek models in the text domain on MuCR.}
\vspace{-0.3cm}
   \label{Sheet2}
\end{table}

\section{Experiments}
In this section, we delve into extended experiments and provide supplementary details that were not included in the main paper for the sake of clarity and brevity.

\subsection{Experimental Results}
\label{A2.1}
As discussed in Section~\ref{M4.1}, we did not include the currently popular models DeepSeek-R1~\cite{guo2025deepseek} and DeepSeek-V3~\cite{liu2024deepseek} in the main paper. Here, we provide a brief comparison of their text-based performance against GPT-o1~\cite{gpto1}. Table~\ref{Sheet2} shows that DeepSeek-R1 achieves results comparable to GPT-o1 in the text domain, while DeepSeek-V3 performs slightly less effectively.

In addition, we provide a detailed breakdown of each model's performance on our MuCR benchmark. Table~\ref{Sheet3} presents these results. We observe that all popular MLLMs significantly outperform random chance, whereas most lightweight open-source models perform below the random baseline of 25\%. This indicates that the latter group lacks robust causal reasoning capabilities.

\section{Cross-modal Generalization Analysis and Enhancement}
\subsection{Picture Style}
\label{A3.1}
\begin{figure*}
    \centering
    \includegraphics[width=0.7\textwidth]{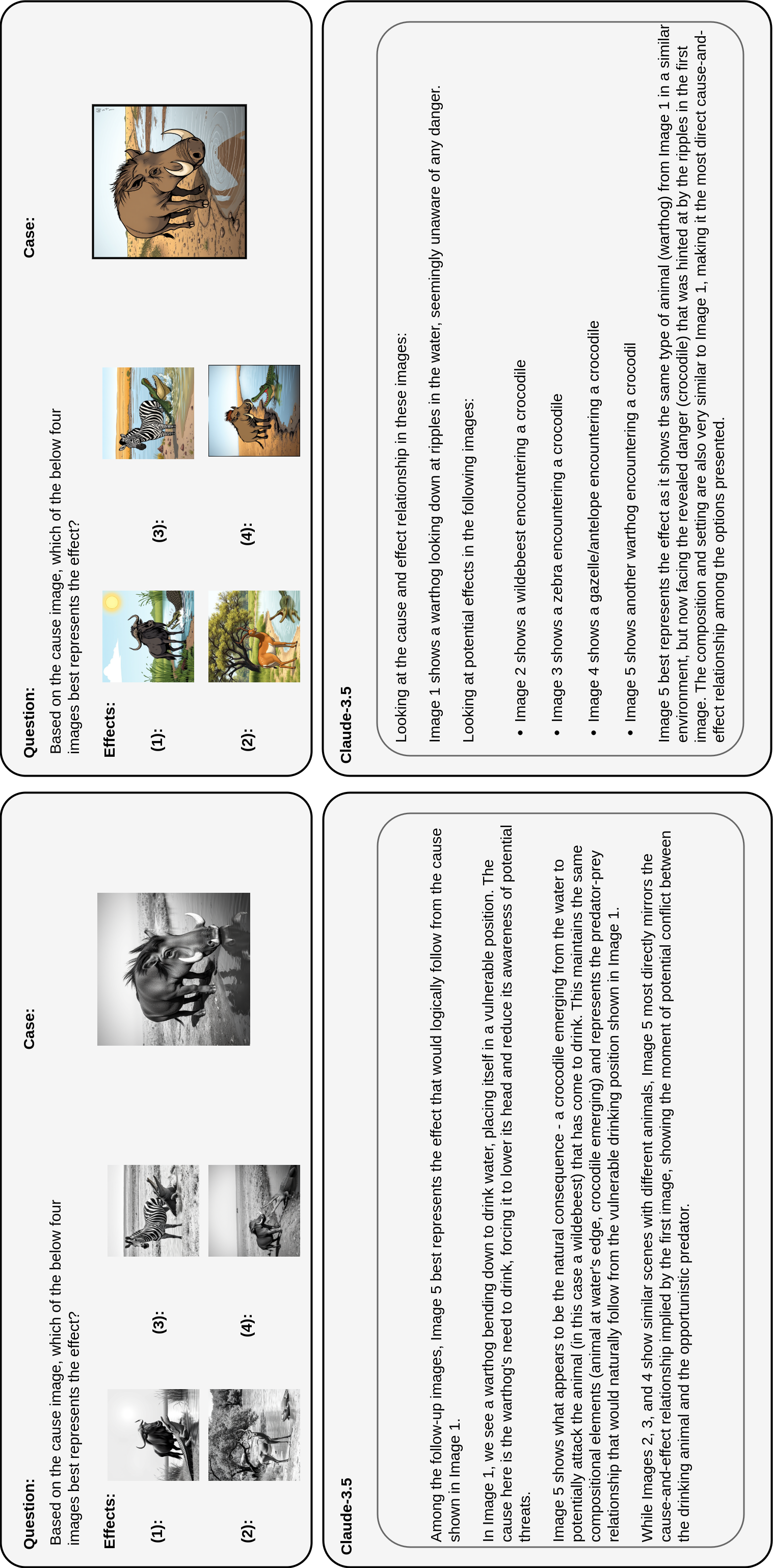}
    \caption{Case study for picture style influence. Best viewed by zooming in.}
    \label{img7}
\end{figure*}

Here, we present a detailed case analysis comparing the influence of picture style on Claude-3.5's predictions, as illustrated in Figure~\ref{img7}. 

In the black-white images, Image 1 shows a warthog bending down to drink water, placing it in a vulnerable position. The cause is clear—the warthog’s need to drink compels it to lower its head, thus reducing its awareness of potential threats. Among the follow-up images, Image 5 best represents the effect: it shows a crocodile emerging from the water, poised to attack a drinking animal, maintaining consistent compositional elements such as the animal at the water’s edge and the predator’s emergence. While Images 2, 3, and 4 depict similar scenarios with different animals, Image 5 most directly mirrors the cause-and-effect relationship suggested by Image 1. However, the analysis in this style tends to lack detail in some of the incorrect answers, which could potentially influence the model’s predictive accuracy in nuanced cases.

In contrast, the comic style analysis also begins with Image 1, where a warthog is depicted looking down at ripples in the water, seemingly unaware of any lurking danger. The potential effects are illustrated across multiple images: Image 2 shows a wildebeest encountering a crocodile, Image 3 depicts a zebra facing a crocodile, Image 4 features a gazelle or antelope in a similar scenario, and Image 5 shows another warthog confronting a crocodile. Here, Image 5 stands out as the best representation of the effect because it features the same animal as in the cause image in a comparable setting, now facing the implied threat signaled by the ripples. The consistent composition and environmental context reinforce the direct cause-and-effect relationship.

The comic style analysis provides a richer context and more detailed narrative for the causal relationship, whereas the balck-white analysis, although accurate in identifying the correct image, offers less detailed reasoning for some incorrect options.

\subsection{Form of Visual Input}
\label{A3.2}

\begin{figure*}
    \centering
    \includegraphics[width=0.8\textwidth]{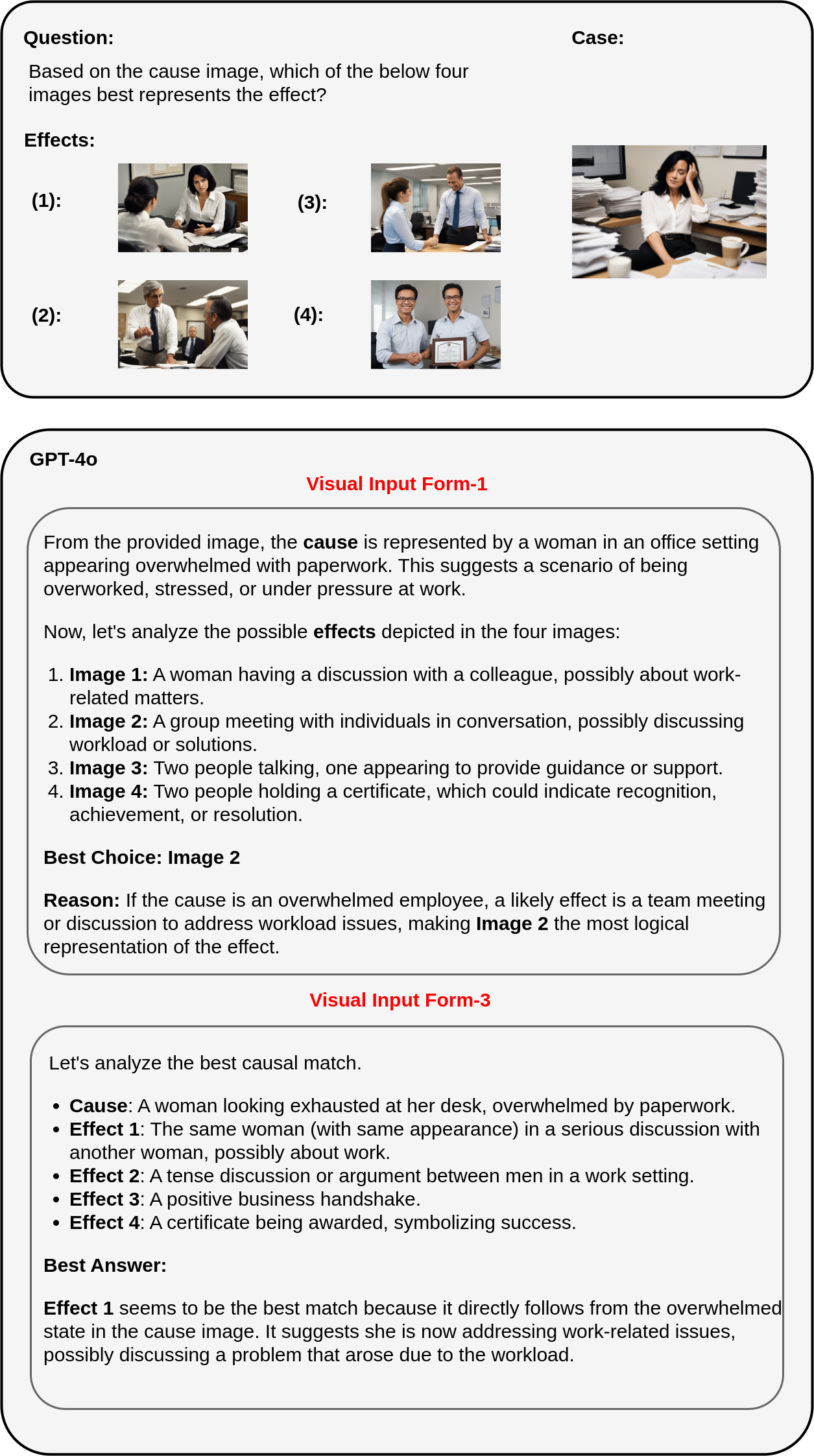}
    \caption{Case study for visual input form influence.}
    \label{img6}
\end{figure*}

Our case analysis demonstrates that, compared to Form-3, Form-1 and Form-2 impose limitations on MLLMs’ ability to recognize and leverage critical visual cues necessary for multimodal causal reasoning. As shown in Figure~\ref{img6}, Form-3 provides GPT-4o with direct visual information, enabling it to successfully identify essential details, such as the continuity in a person’s appearance across cause-and-effect images. This was evident in GPT-4o’s output, where it correctly determined that the woman in the cause image, overwhelmed by paperwork, was the same individual in the effect image, now engaged in a serious discussion about work. This recognition of visual consistency is crucial for establishing causal relationships. However, when using Form-1, GPT-4o was unable to incorporate this specific visual cue and instead selected a different effect image (a generic team meeting) based on a more abstract textual interpretation rather than a direct visual correlation.

The key issue with Form-1 and Form-2 is that they rely on structured textual descriptions that predefine categories of reasoning, which may inadvertently filter out implicit but important visual details. These formats encourage MLLMs to focus on generalized textual patterns rather than independently deriving causal relationships from visual features like facial expressions, body language, and scene continuity. By contrast, Form-3 allows GPT-4o to analyze raw visual inputs more freely, enhancing its ability to establish causal links based on direct visual observation rather than abstracted textual hints. This distinction highlights the potential shortcomings of rigid textual input structures in multimodal causal reasoning tasks. While textual guidance can be helpful, it may also constrain the model’s reasoning process, making it less sensitive to nuanced visual cues. Ensuring that MLLMs receive input formats that preserve rich visual information is therefore essential for improving their ability to perform causal inference in multimodal settings.

\subsection{Contextual Variation}
\label{A3.3}
\begin{figure*}
    \centering
    \includegraphics[width=0.8\textwidth]{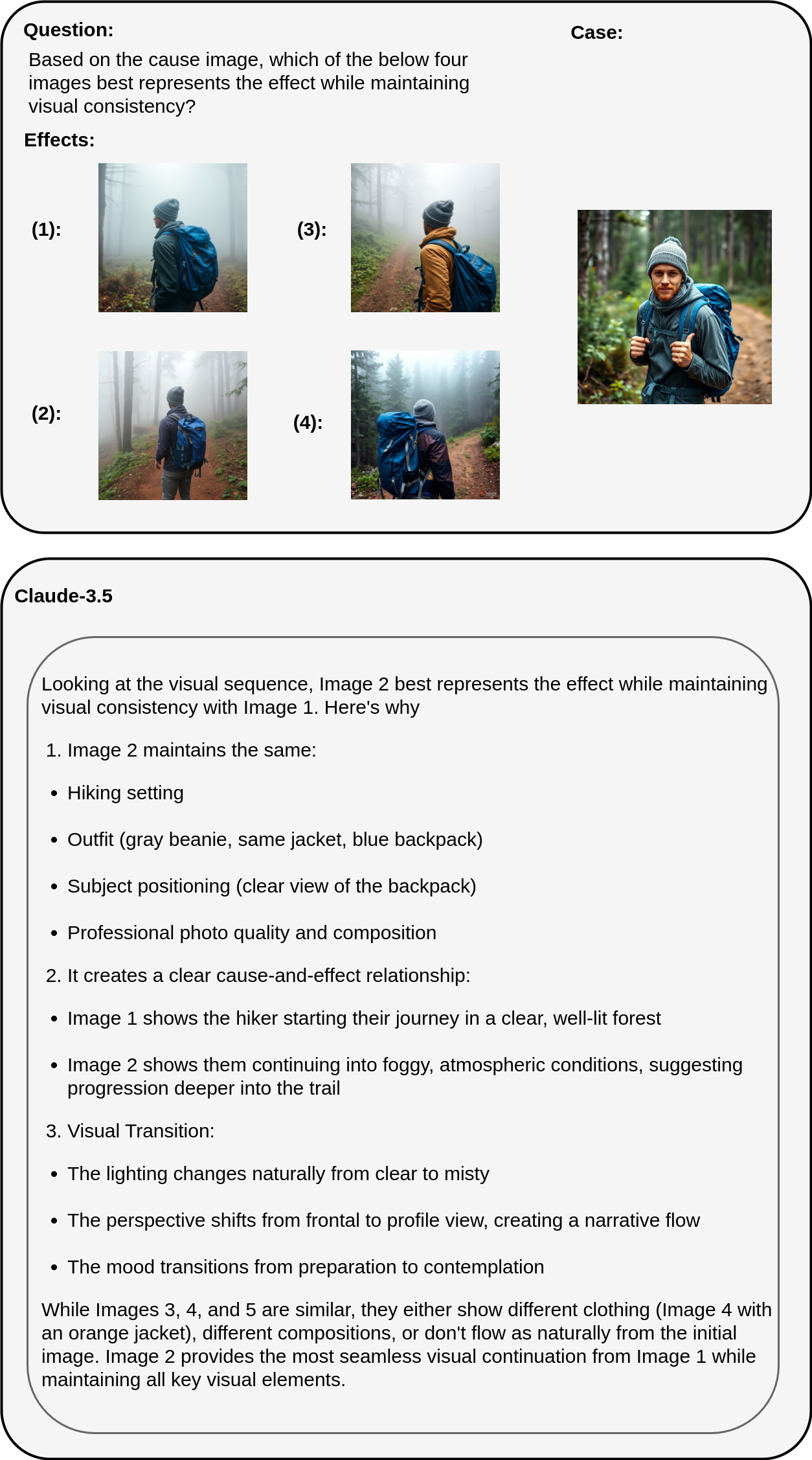}
    \caption{Case study for Contextual Variation.}
    \label{img8}
\end{figure*}

Visual cues are crucial for accurate multimodal causal inference because they provide a consistent framework for linking cause and effect. Taking Figure~\ref{img8} as an example: 
\begin{itemize}
    \item \textbf{Consistency:} Shared elements like the hiking setting, outfit, and subject positioning help the model recognize that the images belong to the same sequence.
    \item \textbf{ Clear Transitions:} Changes in lighting, perspective, and mood signal the progression from cause (a clear, well-lit forest) to effect (a foggy, atmospheric scene), reinforcing the narrative flow.
    \item \textbf{Disambiguation:} Detailed cues identify Image 2 as the best continuation among similar options, ensuring the causal relationship is accurately maintained.
\end{itemize}
The analysis shows that visual cues—ranging from consistent environmental context and subject details to nuanced transitions in lighting, perspective, and mood—are crucial for establishing a clear and coherent narrative. These cues allow the model to accurately determine the causal links between images, ensuring that the inferred relationships are both logical and contextually grounded. Without such detailed visual information, the model would face challenges in differentiating between similar scenarios, potentially leading to inaccurate or incomplete causal inferences.

\subsection{Qualitative Results of VcCOT}
\label{A3.5}
\begin{figure*}
    \centering
    \includegraphics[width=0.75\textwidth]{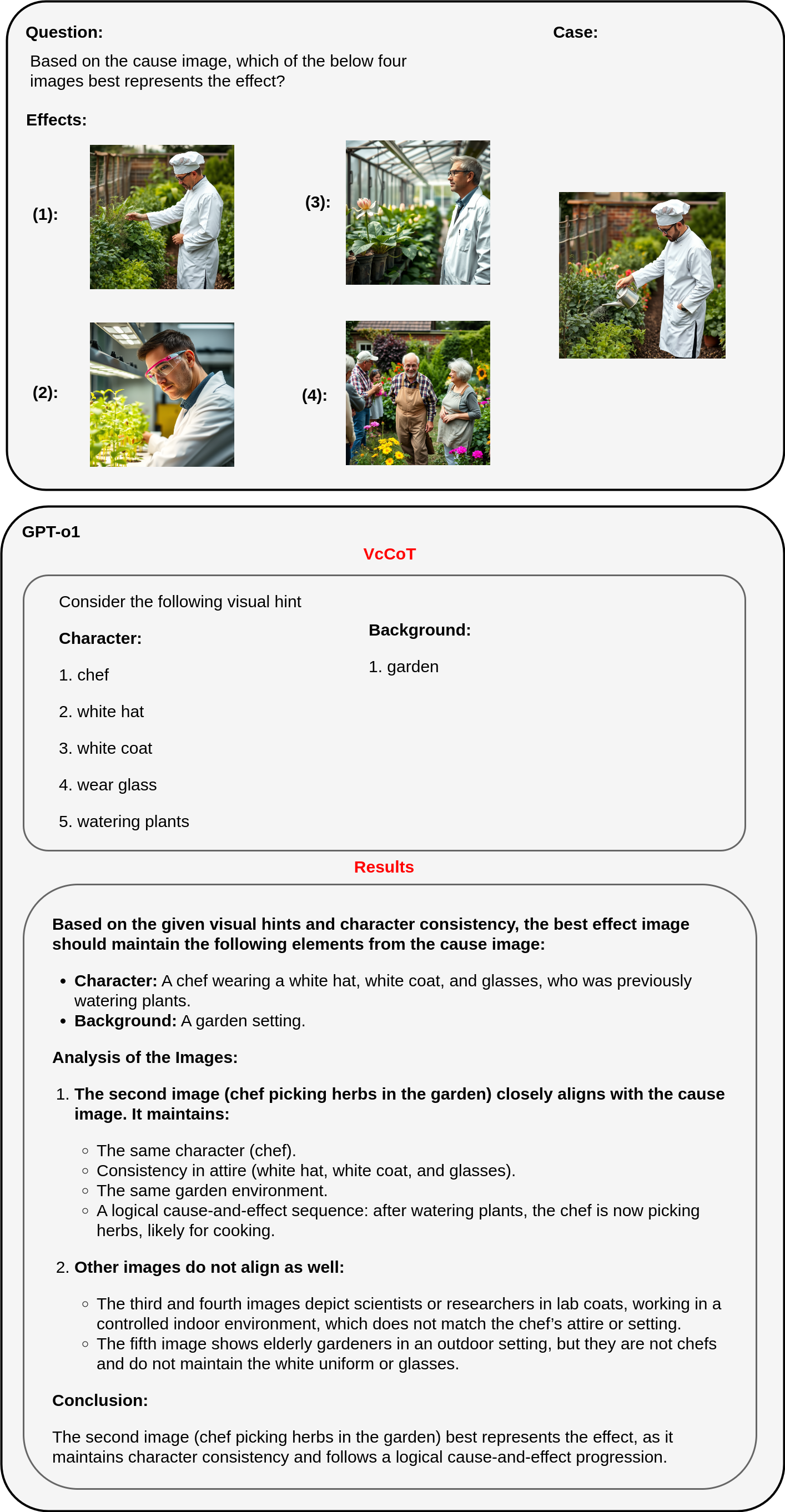}
    \caption{Qualitative results for VcCoT.}
    \label{img9}
\end{figure*}

To prove the efficiency of our VcCoT, we provide some qualitative results, as shown in Figure~\ref{img9}.

\end{document}